\documentclass[10pt,twocolumn,letterpaper]{article}

\usepackage[pagenumbers]{cvpr}      

\usepackage{algorithm}
\usepackage{algpseudocode}
\usepackage{multirow}
\usepackage{tabularx}
\let\REQUIRE\Require
\let\ENSURE\Ensure
\let\STATE\State

\definecolor{cvprblue}{rgb}{0.21,0.49,0.74}
\usepackage[pagebackref,breaklinks,colorlinks,allcolors=cvprblue]{hyperref}

\def\paperID{*****} 
\def\confName{CVPR}
\def\confYear{2026}

\title{RFMSR: Residual Flow Matching for Image Super-Resolution}

\author{Shuwei Huang$^{*}$ \quad Tianyao Luo$^{*}$ \quad Jicheng Liu$^{\dagger}$\\[0.5em]
Huazhong University of Science and Technology\\[0.5em]
$^{*}$Equal contribution \quad $^{\dagger}$Corresponding author
}

\begin{document}
\maketitle
\begin{abstract}
Generative image super-resolution (ISR) methods typically sample from a standard Gaussian prior and transport the sample to the data distribution, where the low-quality (LQ) input serves only as an external condition. 
However, the LQ image already carries rich structural information about the target high-quality (HQ) image, which makes a distribution approximating the LQ image a more suitable starting point, an idea that has been validated by a series of diffusion-based restoration methods. 
In contrast, recent flow-matching-based ISR methods still adopt the standard Gaussian prior and thereby discard such structural priors; moreover, existing single-step acceleration techniques, such as distillation and consistency training, either depend on an additional teacher or sacrifice the model's multi-step inference capability. 
In this paper, we propose \textbf{R}esidual \textbf{F}low \textbf{M}atching for \textbf{I}mage \textbf{S}uper-\textbf{R}esolution (RFMSR), which defines the source distribution as a Gaussian centered at the LQ latent, so that the flow starts from the LQ image rather than from pure noise and preserves the structural priors already present in the input. 
To achieve high-quality single-step generation, we further propose a simple yet effective training strategy that combines the supervision loss on the single-step output with the conditional flow matching (CFM) loss, enabling high-quality restoration with an arbitrary number of steps without resorting to distillation or other acceleration techniques. 
Extensive experiments show that RFMSR achieves competitive results against state-of-the-art methods.
The source code is available at \url{https://github.com/Faze-Hsw/RFMSR}.
\end{abstract}
    
\section{Introduction}
\label{sec:intro}

Image super-resolution (ISR) aims to reconstruct a high-quality (HQ) image from its low-quality (LQ) counterpart. 
While early deep learning methods~\cite{dong2014learning,lim2017enhanced,chen2023activating,liang2021swinir} achieved impressive pixel-level accuracy, they often produce over-smoothed results that lack high-frequency details. 
To improve perceptual quality, GAN-based approaches~\cite{ledig2017photo,wang2018esrgan,wang2021real,zhang2021designing} employ adversarial training to generate realistic textures.
However, GAN training can be unstable and may introduce unpleasant artifacts.
Recent advances in generative modeling, particularly diffusion models~\cite{ho2020denoising,song2020score,song2020denoising} and flow matching~\cite{lipman2022flow,liu2022flow}, have enabled more stable and perceptually rich image restoration by learning to sample from the natural image distribution.

Generative models typically synthesize images by transporting a sample from a standard Gaussian prior $p_1 = \mathcal{N}(0, I)$ to the data distribution. 
For image super-resolution, however, such an uninformative prior is unnecessary: the LQ image already carries rich structural information about the target HQ image, so a distribution that approximates the LQ image is a much more suitable starting point. 
This idea has been successfully exploited by diffusion-based ISR methods. 
ResShift~\cite{yue2023resshift} shortens the diffusion chain by shifting the residual between the LQ and HQ images, IR-SDE~\cite{luo2023image} defines a mean-reverting SDE whose reverse process starts from a Gaussian centered at the LQ image, I$^2$SB~\cite{liu2023i2sb} learns a Schr\"odinger bridge directly between the LQ and HQ distributions, and InDI~\cite{delbracio2023inversion} restores the LQ input through progressive iterative refinement. 
These studies consistently indicate that injecting the LQ prior into the source distribution, rather than relying on conditioning alone, preserves the structures already present in the input and improves the quality of the restored images.

More recently, flow matching~\cite{lipman2022flow,liu2022flow} has emerged as another mainstream generative paradigm owing to its conceptual simplicity and efficient ODE-based sampling, and has been quickly adopted for ISR. 
In particular, a series of flow-matching methods has been developed for ISR and image restoration, such as VOSR~\cite{wu2026vosr}, OMGSR~\cite{wu2025omgsr}, FlowIE~\cite{zhu2024flowie}, and FlowDPS~\cite{kim2025flowdps}. 
Nevertheless, these methods still adopt the standard Gaussian distribution as the source of the flow: the LQ image enters only as a conditioning signal, so the structural prior it carries has to be re-synthesized from noise along the entire trajectory.

\begin{figure}[t]
    \centering
    \includegraphics[width=\linewidth]{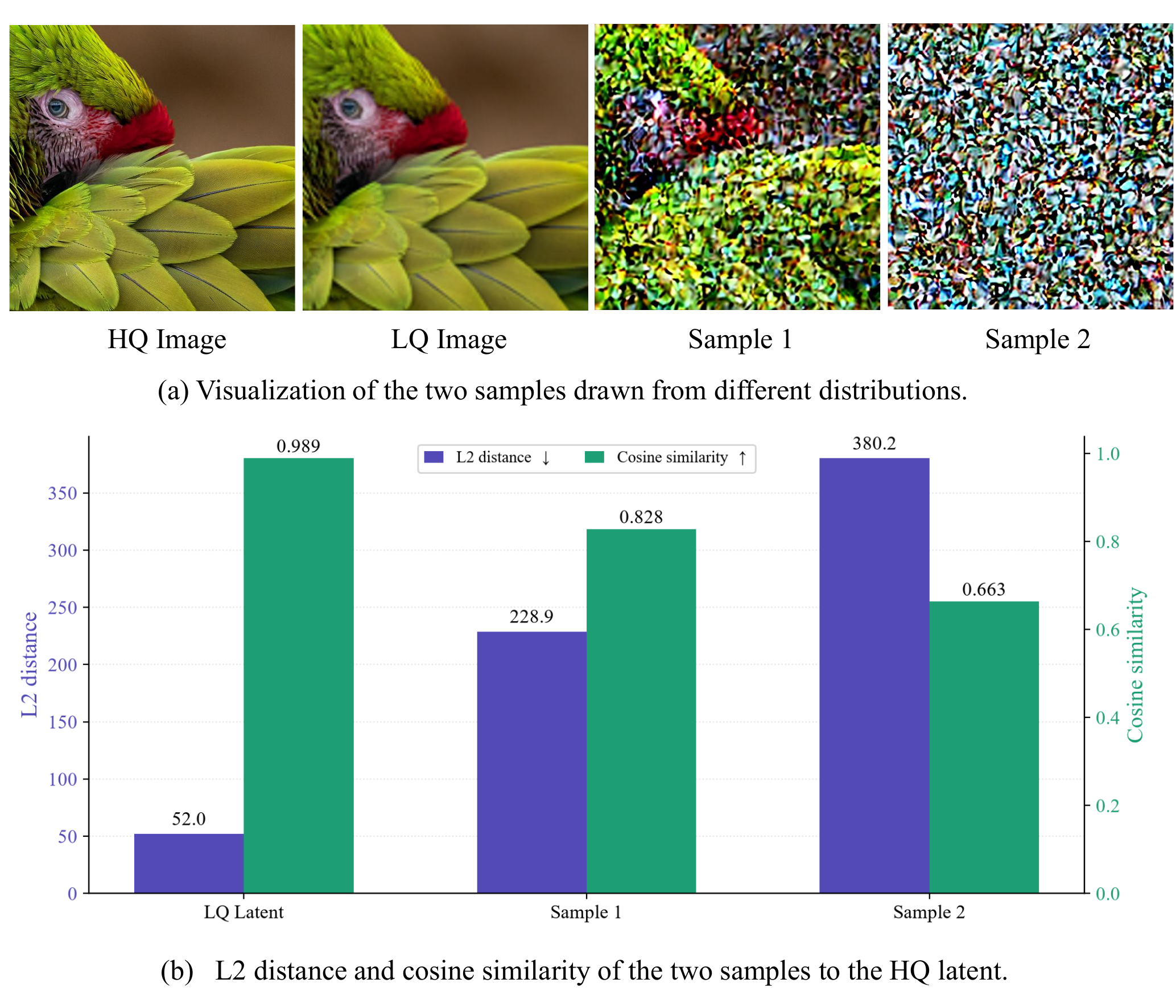}
    \caption{Sample 1 is drawn from $\mathcal{N}(y_0, I)$, while Sample 2 is drawn from $\mathcal{N}(0, I)$.}
    \label{fig:distance_comparison}
\end{figure}

In this paper, we propose \textbf{R}esidual \textbf{F}low \textbf{M}atching for \textbf{I}mage \textbf{S}uper-\textbf{R}esolution (RFMSR), a framework that starts the flow from an LQ-centered distribution instead of pure noise. 
Specifically, we define the source distribution as a Gaussian centered at the LQ latent $y_0$ with variance $\sigma^2$, \ie, $p_1 = \mathcal{N}(y_0, \sigma^2 I)$. 
As shown in \cref{fig:distance_comparison}, an LQ-based source distribution retains richer structural priors than the standard Gaussian distribution, which helps improve the quality of the restored images (see \cref{sec:ablation_studies}).

Beyond architectural design, real-world deployment typically demands single-step generation for efficiency. 
Existing solutions can be roughly grouped into three families: distillation-based methods that compress a pre-trained multi-step model into a one-step student, \eg, the variational score distillation~\cite{wang2023prolificdreamer} adopted by OSEDiff~\cite{wu2024one} and the trajectory distillation of SinSR~\cite{wang2024sinsr}; consistency-based methods that learn a mapping from any point of the sampling trajectory to its endpoint, so that a single evaluation suffices~\cite{song2023consistency}; and shortcut-based methods that condition the network on the desired step size so that it can jump ahead by an arbitrary interval~\cite{frans2025shortcut}. 
In contrast to these methods, we propose a simpler, more direct, and effective alternative for single-step acceleration. 
Our motivation is straightforward: if one-step inference is what deployment requires, the reconstruction objective can be imposed directly on the one-step output $\hat{x}_0$, jointly with the conditional flow matching (CFM) velocity loss over the whole trajectory. 
In this way, a single set of weights delivers high-quality results in both the single-step and the multi-step regimes (see \cref{sec:ablation_studies}).

Our contributions are summarized as follows:

\begin{itemize}
    \item We propose Residual Flow Matching for ISR, which centers the source distribution at the LQ latent and preserves structural priors, improving the quality of the restored images compared with standard rectified flow.
    \item We design a two-phase training strategy that achieves high-quality single-step generation while fully retaining multi-step inference capability, combining the flexibility of flow matching with the efficiency of end-to-end supervision.
    \item Extensive experiments validate the effectiveness of the proposed design and show that our method achieves competitive results compared with state-of-the-art approaches.
\end{itemize}
\section{Related Work}
\label{sec:related_work}

\noindent\textbf{Generative Model for Image Super-Resolution.}
Generative ISR methods build on diffusion models~\cite{ho2020denoising,song2020score} or flow matching~\cite{lipman2022flow,liu2022flow}, typically operating in the latent space of a pre-trained VAE~\cite{rombach2022high}.
Diffusion-based methods have been developed both with large text-to-image (T2I) priors such as Stable Diffusion~\cite{stabilityai}, which are adapted to ISR by adding extra modules or concatenating the LQ image into the input through fine-tuning or distillation, \eg, StableSR~\cite{wang2024exploiting}, DiffBIR~\cite{lin2024diffbir}, SeeSR~\cite{wu2024seesr}, InvSR~\cite{yue2025arbitrary}, PASD~\cite{yang2024pixel}, and SUPIR~\cite{yu2024scaling}, and with vision-only backbones such as SR3~\cite{saharia2022image}, ResShift~\cite{yue2023resshift}, and SinSR~\cite{wang2024sinsr}, which train a native ISR model from scratch.
Notably, a line of these works departs from the standard Gaussian prior and instead starts from a distribution derived from the LQ image: ResShift~\cite{yue2023resshift} shifts the residual between the LQ and HQ images over a shortened Markov chain, IR-SDE~\cite{luo2023image} defines a mean-reverting SDE whose reverse process starts from a Gaussian centered at the LQ image, I$^2$SB~\cite{liu2023i2sb} learns a Schr\"odinger bridge between the LQ and HQ distributions, and InDI~\cite{delbracio2023inversion} progressively refines the LQ input.

More recently, flow matching has been adopted for ISR and image restoration owing to its simple formulation and efficient ODE sampling. 
VOSR~\cite{wu2026vosr} integrates flow matching with DINOv2~\cite{oquab2023dinov2} semantic features and partial conditioning to build a generative ISR model competitive with T2I-based methods. 
OMGSR~\cite{wu2025omgsr} injects the LQ latent at an intermediate timestep to obtain a one-step Real-ISR model. 
FlowIE~\cite{zhu2024flowie} applies a conditioned rectified flow to general image enhancement and straightens the sampling path so that fewer than five steps are sufficient. 
FlowDPS~\cite{kim2025flowdps} solves inverse problems including SR by training-free posterior sampling with pre-trained flow models.
Nevertheless, these flow-matching methods still transport from the standard Gaussian distribution to the HQ latent, so the structural priors carried by the LQ image are injected only as external guidance.

\noindent\textbf{Single-Step Acceleration.}
Since multi-step sampling incurs high inference latency, a series of single-step methods has been developed. 
Most of them adopt a distillation paradigm that transfers a multi-step teacher to a one-step student: OSEDiff~\cite{wu2024one} applies variational score distillation, TSD-SR~\cite{dong2025tsd} introduces target score distillation with real HQ references, YONOS-SR~\cite{noroozi2024you} progressively distills across upscaling factors, SinSR~\cite{wang2024sinsr} distills ResShift along its shortened Markov chain, and AdcSR~\cite{chen2025adversarial} removes redundant modules of a diffusion-based SR model and trains the compressed network adversarially to obtain a compact one-step model. The main drawback of distillation is that the student model is inherently limited by the prediction errors of the teacher. 
A second line removes the teacher by learning to reach the trajectory endpoint within the training objective itself, as in consistency models~\cite{song2023consistency} and shortcut models~\cite{frans2025shortcut}. 
We instead adopt a simpler and more intuitive route to single-step acceleration: directly supervising the one-step prediction effectively improves the single-step generation quality (see \cref{sec:experiments}).

\section{Method}
\label{sec:method}

\begin{figure*}[t]
    \centering
    \includegraphics[width=\linewidth]{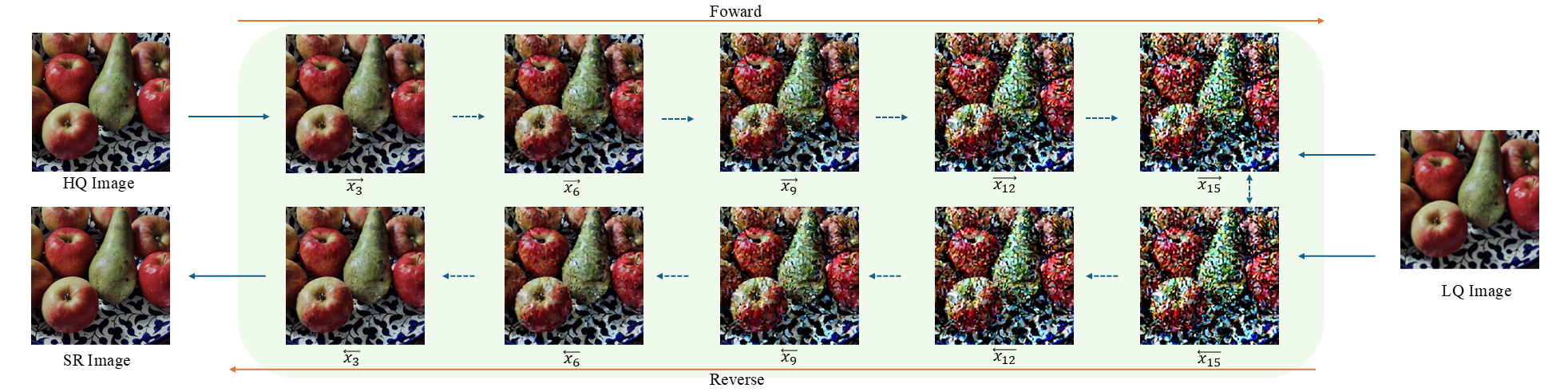}
    \caption{Overview of RFMSR. We uniformly sample 15 points in $[0,1]$ and display a subset of the intermediate states. \textbf{Forward} (top): the HQ latent $x_0$ is progressively transformed toward a Gaussian distribution centered at the LQ latent $y_0$. \textbf{Reverse} (bottom): a latent sampled near $y_0$ is progressively refined into the final ISR result.}
    \label{fig:overview}
\end{figure*}

\begin{figure}[t]
    \centering
    \includegraphics[width=\linewidth]{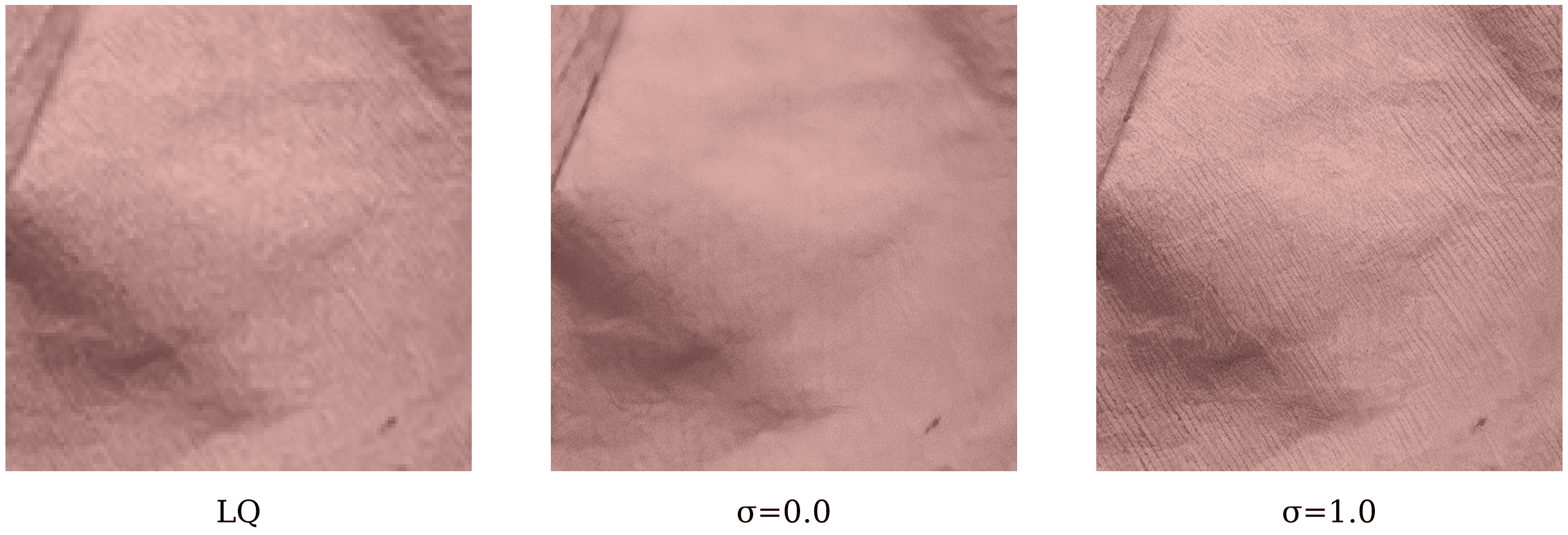} \\[4pt]
    \includegraphics[width=\linewidth]{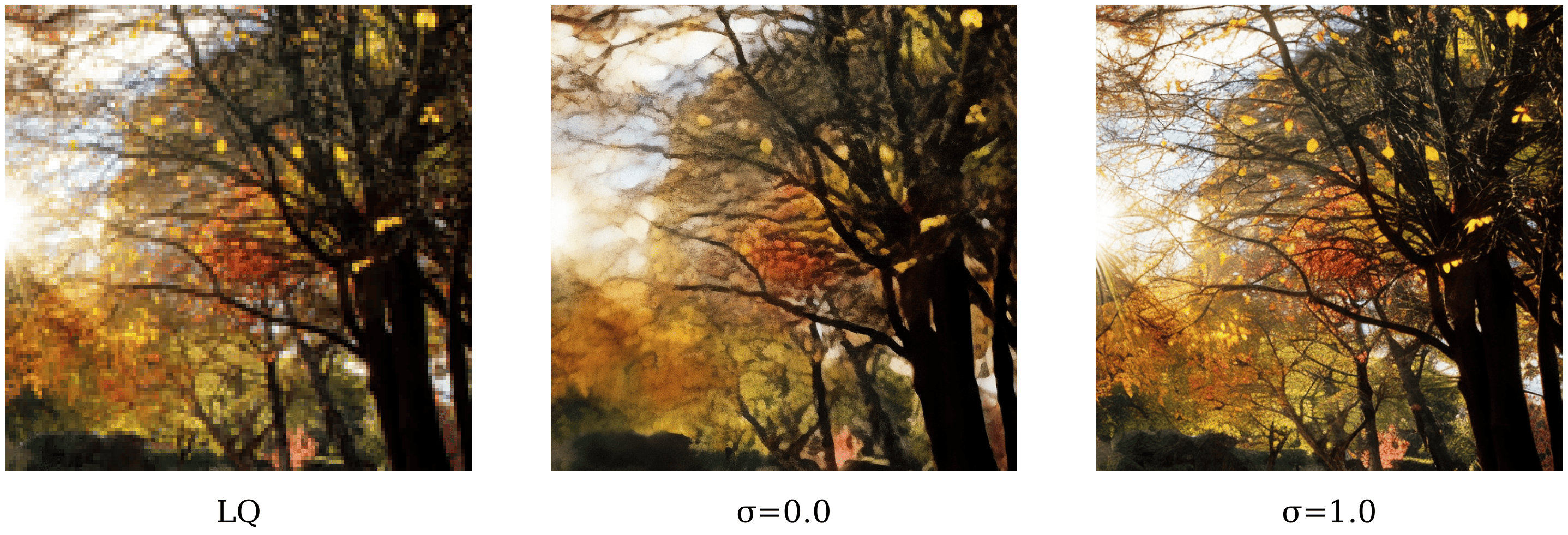}
    \caption{Visual effect of the noise level $\sigma$. At $\sigma=0.0$, the model collapses to a deterministic regression with over-smoothed outputs; increasing $\sigma$ to $1.0$ restores perceptually richer details. (Zoom in for best view.)}
    \label{fig:sigma_ablation}
\end{figure}

\subsection{Preliminaries}
We begin by briefly reviewing the flow matching framework~\cite{lipman2022flow,liu2022flow}. 
Let $p_1$ denote a simple prior distribution (\eg, a Gaussian distribution) and $p_0$ the target data distribution. 
A flow $\phi_t$ governed by the ODE
\begin{equation}
    \frac{dx_t}{dt} = v_\theta(x_t, t), \quad t \in [0, 1],
\end{equation}
transports samples from $p_1$ to $p_0$, where $v_\theta$ is a velocity field parameterized by a neural network. 
Given $x_1 \sim p_1$, a data point $x_0 \sim p_0$ can be generated by integrating $v_\theta$ from $t = 1$ to $t = 0$. This defines a probabilistic path $p_t$ from $p_0$ to $p_1$, where $p_t$ denotes the marginal distribution of $x_t$ at time $t \in [0, 1]$.
Training $v_\theta$ via maximum likelihood in the continuous normalizing flow framework is prohibitively expensive, as the marginal probability path $p_t(x)$ is intractable. Conditional Flow Matching (CFM)~\cite{lipman2022flow} circumvents this by conditioning on both endpoints. 
Specifically, one defines a conditional probability path $p_t(x \mid x_1, x_0)$ and the corresponding conditional velocity field $v_t(x \mid x_1, x_0)$, then minimizes
\begin{equation}
    \mathcal{L}_{\text{CFM}}(\theta) = \mathbb{E}_{t,\, x_1 \sim p_1,\, x_0 \sim p_0}\Big[ \big\| v_\theta(x_t, t) - v_t(x_t \mid x_1, x_0) \big\|^2 \Big],
\end{equation}
where $t \sim \mathcal{U}[0, 1]$. A canonical choice is the straight path proposed by Rectified Flow~\cite{liu2022flow}, which yields
\begin{equation}
    x_t = (1 - t)\, x_0 + t\, x_1, \qquad v_t(x_t \mid x_1, x_0) = x_1 - x_0.
\end{equation}
During inference, one draws $x_1 \sim p_1$ and solves the ODE backward from $t = 1$ to $t = 0$ using a numerical integrator (\eg, Euler method).

\subsection{Residual Flow Matching for ISR}
The standard flow matching formulation assumes $p_1$ to be a standard Gaussian $\mathcal{N}(0, I)$, which is a natural choice for T2I generation tasks~\cite{liu2023instaflow,esser2024scaling} where the target distribution is conditioned on a text prompt rather than an input image.
For ISR tasks, however, the LQ image provides substantial structural information about the target HQ image. Starting from pure noise discards this information, forcing the model to re-synthesize the structures already present in the LQ image from scratch.

Following~\cite{yue2023resshift}, we propose Residual Flow Matching for ISR (RFMSR), which constructs the source distribution centered around the LQ image.
Let $x_0, y_0$ denote the HQ and LQ images encoded by a pre-trained VAE encoder (the SD2.1 VAE~\cite{podell2024sdxl} for the 0.5B variant and a 16-channel Qwen-Image VAE~\cite{wu2025qwenimagetechnicalreport} for the 1.4B variant in our implementation). 
We define the source distribution as
\begin{equation}
    p_1 = \mathcal{N}\big(y_0,\, \sigma^2 I\big),
\end{equation}
where $\sigma$ is a hyperparameter that controls the initial noise strength. The conditional flow path from source ($t = 1$) to target ($t = 0$) is a linear interpolation augmented with a time-dependent noise injection:
\begin{equation}
    x_t = x_0 + t \cdot \big(y_0 - x_0\big) + t \cdot \sigma \cdot \varepsilon, \quad \varepsilon \sim \mathcal{N}(0, I).
    \label{eq:rfmsr_path}
\end{equation}
Notably, at $t = 0$, $x_t$ converges to the clean HQ latent $x_0$, while at $t = 1$, $x_t$ follows a Gaussian distribution centered at the LQ latent $y_0$.
The conditional velocity field $v_t$ follows directly from differentiation:
\begin{equation}
    v_t(x_t \mid x_1, x_0) = \frac{d x_t}{d t} = x_1 - x_0, \quad x_1 \sim \mathcal{N}(y_0, \sigma^2 I).
\end{equation}
The training loss follows directly from the CFM framework:
\begin{equation}
    \mathcal{L}(\theta) = \mathbb{E}_{t,\, x_0 \sim p_0,\, x_1 \sim \mathcal{N}(y_0, \sigma^2 I)}\Big[ \big\| v_\theta(x_t, t, c) - (x_1 - x_0) \big\|^2 \Big].
    \label{eq:rfm_loss}
\end{equation}
where $c$ denotes the LQ conditioning information.

An illustrative trajectory of the proposed residual flow is shown in \cref{fig:overview}. Notably, the initial state sampled from $p_1$ already retains the structural information of the LQ image, which significantly reduces the difficulty of the restoration process compared to starting from pure noise.

\subsection{Why Noise Matters}
\label{sec:why_noise}

A natural simplification of \cref{eq:rfmsr_path} would be to remove the noise term entirely, \ie, $x_t = x_0 + t\,(y_0 - x_0)$. While conceptually simple, this design suffers from a fundamental limitation, as the resulting deterministic path corresponds to $v_t = y_0 - x_0$ and reduces the learning objective to a plain $\mathcal{L}_2$ regression of the residual.
\\
\textbf{Perceptual degradation.} ISR is an inherently ill-posed, one-to-many inverse problem: a single LQ image admits multiple plausible HQ reconstructions that differ in fine-grained textures and high-frequency details. 
Under the deterministic path, the $\mathcal{L}_2$ objective forces the model to predict the conditional expectation $\mathbb{E}[x_0 \mid y_0]$, which is the pixel-wise mean over all plausible HQ reconstructions. This leads to over-smoothed outputs that lack high-frequency details. 
As shown in \cref{fig:sigma_ablation}, the deterministic variant ($\sigma=0.0$) produces noticeably blurrier results than RFMSR ($\sigma=1.0$), particularly in textured regions. 
Quantitative results in \cref{tab:sigma_ablation} further confirm this, showing a significant degradation in perceptual metrics.

\subsection{Single-Step Generation via Two-Phase Training}

\begin{algorithm}[t]
\caption{Phase~II Training Procedure}
\label{alg:two_phase}
\begin{algorithmic}[1]
\REQUIRE Training dataset $\mathcal{D} = \{(\text{HQ}_i, \text{LQ}_i)\}$, pre-trained VAE $\text{Enc}$, $\text{Dec}$, Phase~I checkpoint, iterations $N$.

\medskip
\STATE Load $\theta$ from the Phase~I checkpoint.
\For{$n \gets 1$ \textbf{to} $N$}
    \STATE Sample $(\text{HQ}, \text{LQ})$ from $\mathcal{D}$, encode $x_0 \leftarrow \text{Enc}(\text{HQ})$, $y_0 \leftarrow \text{Enc}(\text{LQ})$
    \STATE \textbf{(a) CFM velocity loss}
    \STATE Sample $t \sim \mathcal{U}[0, 1]$,  $\varepsilon \sim \mathcal{N}(0, I)$
    \STATE $x_t \leftarrow x_0 + t\,(y_0 - x_0) + t\,\sigma\,\varepsilon$
    \STATE $\mathcal{L} \leftarrow \|v_\theta(x_t, t, c) - (y_0 - x_0 + \sigma\,\varepsilon)\|^2$
    \STATE \textbf{(b) End-to-end losses at $t = 1$}
    \STATE $\varepsilon_1 \sim \mathcal{N}(0, I)$, \quad $x_1 \leftarrow y_0 + \sigma\,\varepsilon_1$
    \STATE $\hat{x}_0 \leftarrow x_1 - v_\theta(x_1, 1, c)$
    \STATE $\mathcal{L}_{\text{LPIPS}} \leftarrow \mathcal{L}_{\text{LPIPS}}(\text{Dec}(\hat{x}_0), \text{HQ})$
    \STATE $\mathcal{L}_{\text{adv}} \leftarrow \text{GAN loss}(\hat{x}_0, x_0)$
    \STATE $\mathcal{L}_{\text{total}} \leftarrow \mathcal{L} + \lambda_{\text{LPIPS}}\mathcal{L}_{\text{LPIPS}} + \lambda_{\text{GAN}}\mathcal{L}_{\text{adv}}$
    \STATE $\theta \leftarrow \theta - \eta_\theta \nabla_\theta \mathcal{L}_{\text{total}}$
\EndFor

\medskip
\ENSURE Trained velocity network $v_\theta$.
\end{algorithmic}
\end{algorithm}

While flow matching models naturally support multi-step inference with arbitrary step counts, real-world deployment often demands single-step generation for low latency. 
Recent works accelerate flow-based or diffusion-based ISR through knowledge distillation~\cite{wu2024one,noroozi2024you,dong2025tsd,li2025one,xu2025fast} from large-scale pre-trained text-to-image (T2I) models to produce a one-step student model. 
However, these approaches typically sacrifice the model's native multi-step capability: the distilled student operates exclusively in the one-step regime and no longer benefits from progressive refinement, and the performance of distilled models is inherently bounded by that of the teacher model.
Moreover, distilling such models also transfers their text-to-image generative prior, which is not grounded in the LQ observation, so the student may synthesize textures that look plausible but are inconsistent with the input.
In contrast, RFMSR is vision-only and involves no T2I prior, so its restoration stays anchored to the LQ input.
We propose a two-phase training strategy that endows the model with high-quality single-step generation while fully preserving its multi-step inference capability.
\\
\textbf{Phase I: Velocity Field Pretraining.} In the first phase, we train the model using only the CFM objective defined in~\cref{eq:rfm_loss}. 
The model learns the conditional velocity field $v_\theta(x_t, t, c)$ across all $t \in [0, 1]$, establishing a smooth transport from the source distribution $\mathcal{N}(y_0, \sigma^2 I)$ to the target HQ latent distribution. 
This phase produces a general-purpose flow model capable of multi-step Euler integration.
\\
\textbf{Phase II: End-to-End Refinement.} Starting from the Phase~I checkpoint, we introduce two end-to-end losses operating on the one-step prediction $\hat{x}_0 = x_1 - v_\theta(x_1, 1, c)$: an LPIPS perceptual loss~\cite{zhang2018unreasonable} and a PatchGAN~\cite{isola2017image} with hinge loss~\cite{miyato2018spectral}.
Crucially, we retain the CFM velocity loss $\mathcal{L}(\theta)$ from Phase~I. The full Phase~II objective is:
\begin{equation}
    \mathcal{L}_{\text{Phase II}} = \mathcal{L}(\theta) + \lambda_{\text{LPIPS}}\, \mathcal{L}_{\text{LPIPS}}(\hat{x}_0,\, x_0) + \lambda_{\text{GAN}}\, \mathcal{L}_{\text{adv}}(\hat{x}_0,\, x_0),
\end{equation}
where the CFM term continues to sample $t \sim \mathcal{U}[0, 1]$.
The Phase~II training procedure is summarized in Algorithm~\ref{alg:two_phase}.

\section{Experiments}
\label{sec:experiments}

\begin{table*}[t]
    \centering
    \caption{Quantitative comparison on LSDIR, ImageNet-Test, and ScreenSR. Methods are grouped into multi-step and single-step settings. The first number following each method name denotes its inference steps, and the trailing 0.5B or 1.4B denotes the model scale of VOSR and RFMSR. $\uparrow$ ($\downarrow$) indicates higher (lower) is better. The best and second-best results are highlighted in \textbf{\textcolor{red}{red}} and \textbf{\textcolor{blue}{blue}}, respectively, within each group.}
    \label{tab:main_comparison}
    \resizebox{\linewidth}{!}{%
    \footnotesize
    \begin{tabular}{@{}cclccccccccc@{}}
        \toprule
        Dataset & Type & Method & PSNR$\uparrow$ & SSIM$\uparrow$ & LPIPS$\downarrow$ & DISTS$\downarrow$ & NIQE$\downarrow$ & MUSIQ$\uparrow$ & MANIQA$\uparrow$ & CLIPIQA$\uparrow$ & FID$\downarrow$ \\
        \midrule
        \multirow{12}{*}{LSDIR} & \multirow{6}{*}{Multi-Steps} & SeeSR-50 & \textbf{\textcolor{blue}{20.27}} & 0.5049 & 0.2538 & 0.1528 & \textbf{\textcolor{blue}{4.1000}} & 72.3832 & \textbf{\textcolor{blue}{0.5594}} & \textbf{\textcolor{blue}{0.7142}} & 50.92 \\
         &  & VOSR-25-0.5B & 20.05 & 0.4972 & 0.2348 & 0.1455 & \textbf{\textcolor{red}{3.8119}} & 71.5792 & 0.4729 & 0.6184 & 45.22 \\
         &  & VOSR-25-1.4B & 20.05 & 0.5069 & 0.2461 & 0.1534 & 4.5866 & \textbf{\textcolor{blue}{72.6935}} & 0.4770 & 0.5948 & 48.71 \\
         &  & ResShift-4 & \textbf{\textcolor{red}{20.64}} & \textbf{\textcolor{red}{0.5277}} & 0.2251 & 0.1422 & 5.3839 & 67.3639 & 0.4360 & 0.6436 & \textbf{\textcolor{blue}{37.54}} \\
         &  & \textbf{RFMSR-15-0.5B} & 20.07 & 0.4914 & \textbf{\textcolor{blue}{0.2127}} & \textbf{\textcolor{blue}{0.1293}} & 4.1295 & 72.1519 & \textbf{\textcolor{red}{0.5792}} & \textbf{\textcolor{red}{0.7428}} & 38.91 \\
         &  & \textbf{RFMSR-15-1.4B} & 19.68 & \textbf{\textcolor{blue}{0.5097}} & \textbf{\textcolor{red}{0.1916}} & \textbf{\textcolor{red}{0.1194}} & 4.6871 & \textbf{\textcolor{red}{73.1526}} & 0.5463 & 0.6917 & \textbf{\textcolor{red}{30.86}} \\
        \cmidrule{2-12}
         & \multirow{6}{*}{Single-Step} & VOSR-1-0.5B & \textbf{\textcolor{blue}{20.04}} & 0.4917 & 0.2448 & 0.1524 & \textbf{\textcolor{red}{3.7786}} & 72.1925 & 0.5025 & 0.6303 & 47.79 \\
         &  & VOSR-1-1.4B & 19.94 & 0.4948 & 0.2524 & 0.1526 & 4.2650 & \textbf{\textcolor{red}{73.2562}} & 0.4732 & 0.5977 & 45.06 \\
         &  & InvSR-1 & 19.19 & 0.4845 & 0.2657 & 0.1636 & 4.2561 & 72.7310 & 0.5113 & \textbf{\textcolor{red}{0.7156}} & 69.30 \\
         &  & OSEDiff-1 & 20.03 & 0.5094 & 0.2675 & 0.1627 & 4.1127 & 71.4740 & 0.4573 & 0.6946 & 58.98 \\
         &  & \textbf{RFMSR-1-0.5B} & \textbf{\textcolor{red}{20.10}} & \textbf{\textcolor{blue}{0.5116}} & \textbf{\textcolor{red}{0.1961}} & \textbf{\textcolor{blue}{0.1257}} & \textbf{\textcolor{blue}{3.9389}} & 71.7196 & \textbf{\textcolor{red}{0.5182}} & \textbf{\textcolor{blue}{0.7111}} & \textbf{\textcolor{blue}{37.87}} \\
         &  & \textbf{RFMSR-1-1.4B} & 19.57 & \textbf{\textcolor{red}{0.5135}} & \textbf{\textcolor{blue}{0.2022}} & \textbf{\textcolor{red}{0.1182}} & 4.8487 & \textbf{\textcolor{blue}{72.7570}} & \textbf{\textcolor{blue}{0.5138}} & 0.6645 & \textbf{\textcolor{red}{37.34}} \\
        \midrule
        \multirow{12}{*}{ImageNet-Test} & \multirow{6}{*}{Multi-Steps} & SeeSR-50 & 25.45 & 0.6931 & 0.2547 & 0.1646 & 4.3292 & 72.0539 & \textbf{\textcolor{red}{0.5481}} & \textbf{\textcolor{blue}{0.7072}} & 50.19 \\
         &  & VOSR-25-0.5B & 25.39 & 0.6866 & 0.2322 & 0.1526 & \textbf{\textcolor{blue}{4.3232}} & 72.1132 & 0.4958 & 0.6254 & 43.13 \\
         &  & VOSR-25-1.4B & 25.50 & 0.7005 & 0.2289 & 0.1559 & 4.7199 & \textbf{\textcolor{blue}{72.6481}} & 0.4719 & 0.6024 & 44.73 \\
         &  & ResShift-4 & \textbf{\textcolor{red}{26.95}} & \textbf{\textcolor{red}{0.7390}} & \textbf{\textcolor{red}{0.2060}} & 0.1502 & 5.7940 & 65.3497 & 0.3935 & 0.6294 & \textbf{\textcolor{blue}{35.96}} \\
         &  & \textbf{RFMSR-15-0.5B} & \textbf{\textcolor{blue}{25.77}} & 0.6881 & 0.2330 & \textbf{\textcolor{red}{0.1472}} & \textbf{\textcolor{red}{4.3128}} & 71.1203 & \textbf{\textcolor{blue}{0.5442}} & \textbf{\textcolor{red}{0.7242}} & 38.41 \\
         &  & \textbf{RFMSR-15-1.4B} & 25.58 & \textbf{\textcolor{blue}{0.7028}} & \textbf{\textcolor{blue}{0.2271}} & \textbf{\textcolor{blue}{0.1485}} & 5.2903 & \textbf{\textcolor{red}{72.7177}} & 0.5174 & 0.6771 & \textbf{\textcolor{red}{34.73}} \\
        \cmidrule{2-12}
         & \multirow{6}{*}{Single-Step} & VOSR-1-0.5B & 24.98 & 0.6721 & 0.2445 & 0.1556 & \textbf{\textcolor{red}{4.0571}} & \textbf{\textcolor{blue}{72.3900}} & \textbf{\textcolor{blue}{0.4913}} & 0.6327 & 45.64 \\
         &  & VOSR-1-1.4B & 24.92 & 0.6786 & 0.2396 & 0.1569 & 4.4653 & \textbf{\textcolor{red}{73.1072}} & 0.4658 & 0.6101 & 45.68 \\
         &  & InvSR-1 & 23.98 & 0.6689 & 0.2527 & 0.1616 & 4.2609 & 71.8737 & 0.4702 & \textbf{\textcolor{red}{0.7152}} & 57.64 \\
         &  & OSEDiff-1 & 24.67 & 0.6917 & 0.2462 & 0.1544 & 4.2453 & 71.6051 & 0.4595 & 0.6789 & 50.32 \\
         &  & \textbf{RFMSR-1-0.5B} & \textbf{\textcolor{red}{26.19}} & \textbf{\textcolor{blue}{0.7197}} & \textbf{\textcolor{blue}{0.2002}} & \textbf{\textcolor{blue}{0.1325}} & \textbf{\textcolor{blue}{4.1559}} & 71.3530 & \textbf{\textcolor{red}{0.5137}} & \textbf{\textcolor{blue}{0.6985}} & \textbf{\textcolor{blue}{36.09}} \\
         &  & \textbf{RFMSR-1-1.4B} & \textbf{\textcolor{blue}{26.10}} & \textbf{\textcolor{red}{0.7371}} & \textbf{\textcolor{red}{0.1794}} & \textbf{\textcolor{red}{0.1238}} & 5.2004 & 71.1927 & 0.4697 & 0.6481 & \textbf{\textcolor{red}{32.90}} \\
        \midrule
        \multirow{12}{*}{ScreenSR} & \multirow{6}{*}{Multi-Steps} & SeeSR-50 & \textbf{\textcolor{blue}{22.45}} & \textbf{\textcolor{blue}{0.6273}} & \textbf{\textcolor{blue}{0.2234}} & \textbf{\textcolor{red}{0.1372}} & \textbf{\textcolor{blue}{4.2005}} & 72.3069 & \textbf{\textcolor{blue}{0.5760}} & \textbf{\textcolor{blue}{0.7077}} & 65.50 \\
         &  & VOSR-25-0.5B & 21.67 & 0.5963 & 0.2399 & 0.1520 & \textbf{\textcolor{red}{4.1519}} & 71.5834 & 0.4869 & 0.6257 & 67.69 \\
         &  & VOSR-25-1.4B & 21.79 & 0.6078 & 0.2427 & \textbf{\textcolor{blue}{0.1482}} & 4.7452 & 72.1709 & 0.4813 & 0.5847 & 70.94 \\
         &  & ResShift-4 & \textbf{\textcolor{red}{23.15}} & \textbf{\textcolor{red}{0.6647}} & \textbf{\textcolor{red}{0.2193}} & 0.1517 & 5.7495 & 68.5746 & 0.4199 & 0.6137 & 61.44 \\
         &  & \textbf{RFMSR-15-0.5B} & 21.71 & 0.5982 & 0.2317 & 0.1509 & 4.7840 & \textbf{\textcolor{blue}{72.3417}} & \textbf{\textcolor{red}{0.5838}} & \textbf{\textcolor{red}{0.7390}} & \textbf{\textcolor{blue}{60.93}} \\
         &  & \textbf{RFMSR-15-1.4B} & 21.82 & 0.6190 & 0.2318 & 0.1633 & 5.6634 & \textbf{\textcolor{red}{73.8101}} & 0.5490 & 0.6617 & \textbf{\textcolor{red}{58.02}} \\
        \cmidrule{2-12}
         & \multirow{6}{*}{Single-Step} & VOSR-1-0.5B & 21.91 & 0.6141 & 0.2142 & 0.1365 & \textbf{\textcolor{red}{4.0346}} & \textbf{\textcolor{blue}{71.9357}} & \textbf{\textcolor{blue}{0.5100}} & 0.6439 & 62.89 \\
         &  & VOSR-1-1.4B & 21.69 & 0.6171 & 0.2183 & 0.1383 & 4.4021 & \textbf{\textcolor{red}{72.9638}} & 0.4842 & 0.6147 & 61.43 \\
         &  & InvSR-1 & 21.30 & 0.6100 & 0.2513 & 0.1584 & 4.1617 & 71.8109 & 0.4786 & \textbf{\textcolor{blue}{0.6939}} & 89.04 \\
         &  & OSEDiff-1 & 22.11 & 0.6303 & 0.2303 & 0.1528 & \textbf{\textcolor{blue}{4.0802}} & 71.4191 & 0.4521 & 0.6757 & 74.74 \\
         &  & \textbf{RFMSR-1-0.5B} & \textbf{\textcolor{red}{22.64}} & \textbf{\textcolor{blue}{0.6587}} & \textbf{\textcolor{red}{0.1871}} & \textbf{\textcolor{blue}{0.1342}} & 4.2051 & 71.9098 & \textbf{\textcolor{red}{0.5300}} & \textbf{\textcolor{red}{0.7212}} & \textbf{\textcolor{red}{59.48}} \\
         &  & \textbf{RFMSR-1-1.4B} & \textbf{\textcolor{blue}{22.51}} & \textbf{\textcolor{red}{0.6780}} & \textbf{\textcolor{blue}{0.1960}} & \textbf{\textcolor{red}{0.1304}} & 5.1076 & 71.7023 & 0.4880 & 0.6436 & \textbf{\textcolor{blue}{60.32}} \\
        \bottomrule
    \end{tabular}%
    }
\end{table*}

\begin{figure*}[t]
    \centering
    \includegraphics[width=\linewidth]{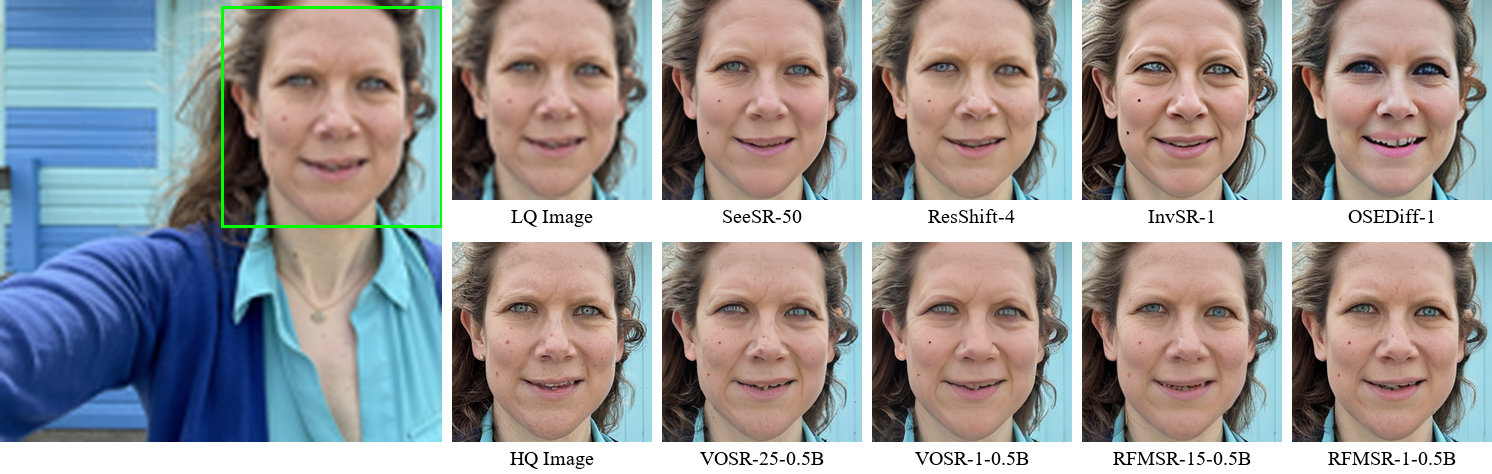} \\[2pt]
    \includegraphics[width=\linewidth]{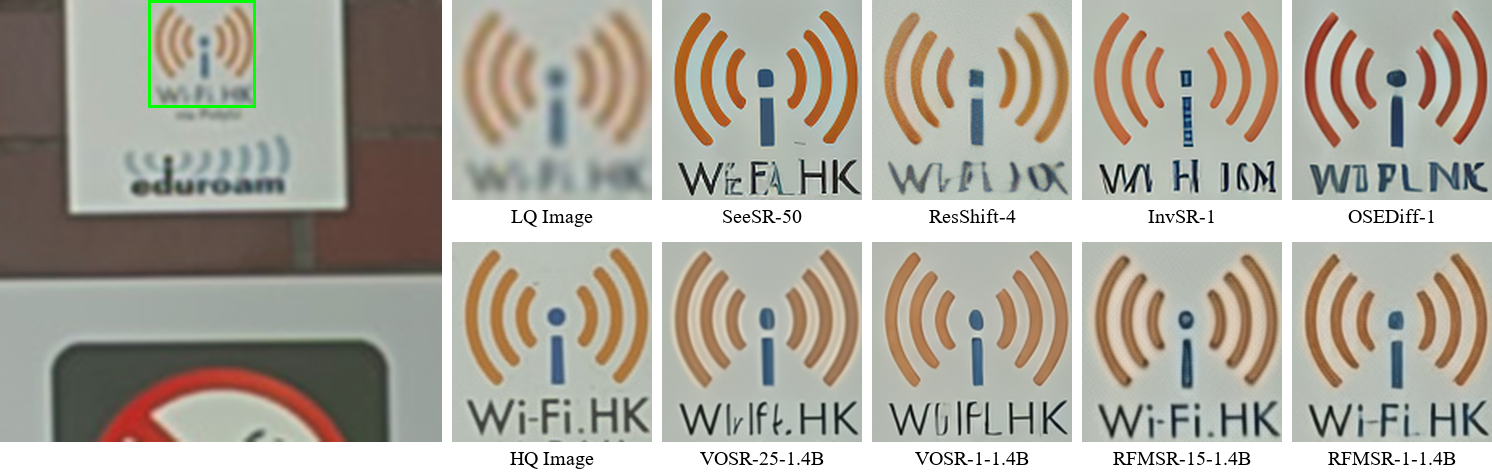}
    \caption{Visual comparison of RFMSR with state-of-the-art methods on real-world images. The first number following each method name denotes its inference steps, and the trailing 0.5B or 1.4B denotes the model scale. (Zoom in for best view.)}
    \label{fig:sr_comparison}
\end{figure*}

\begin{figure}[t]
    \centering
    \includegraphics[width=\linewidth]{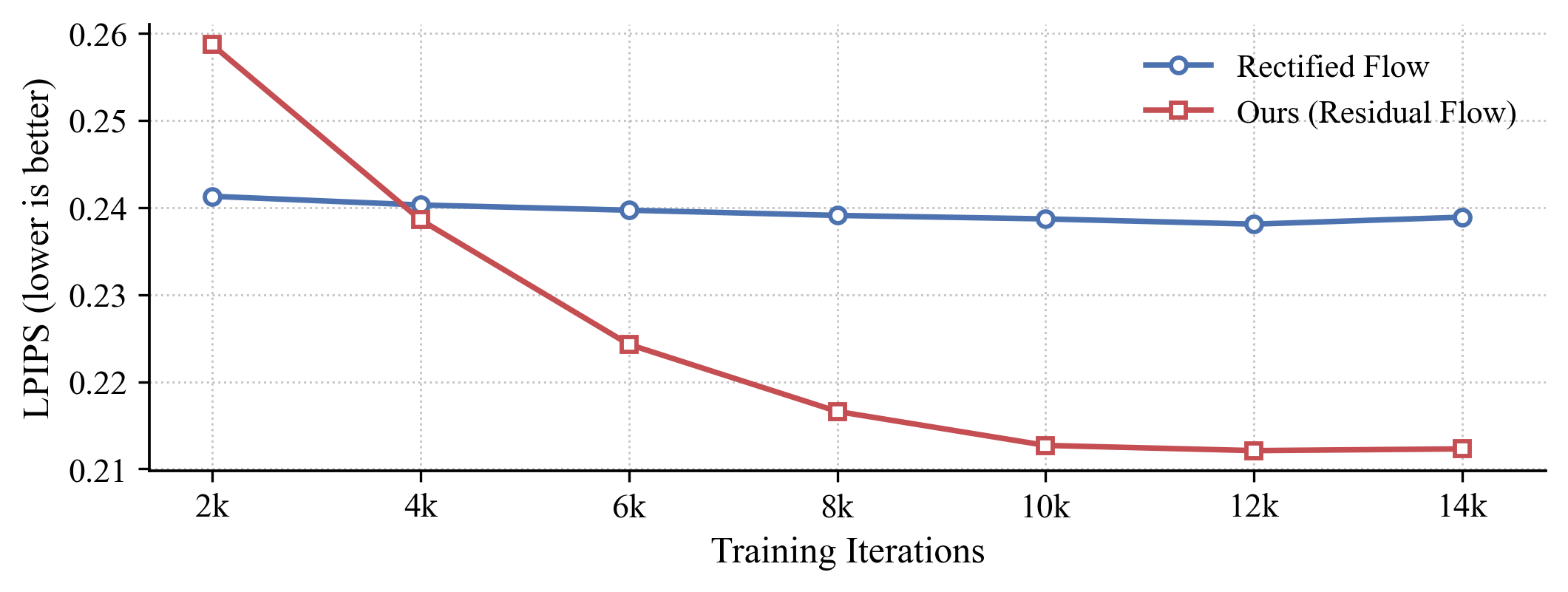} \\[2pt]
    \includegraphics[width=\linewidth]{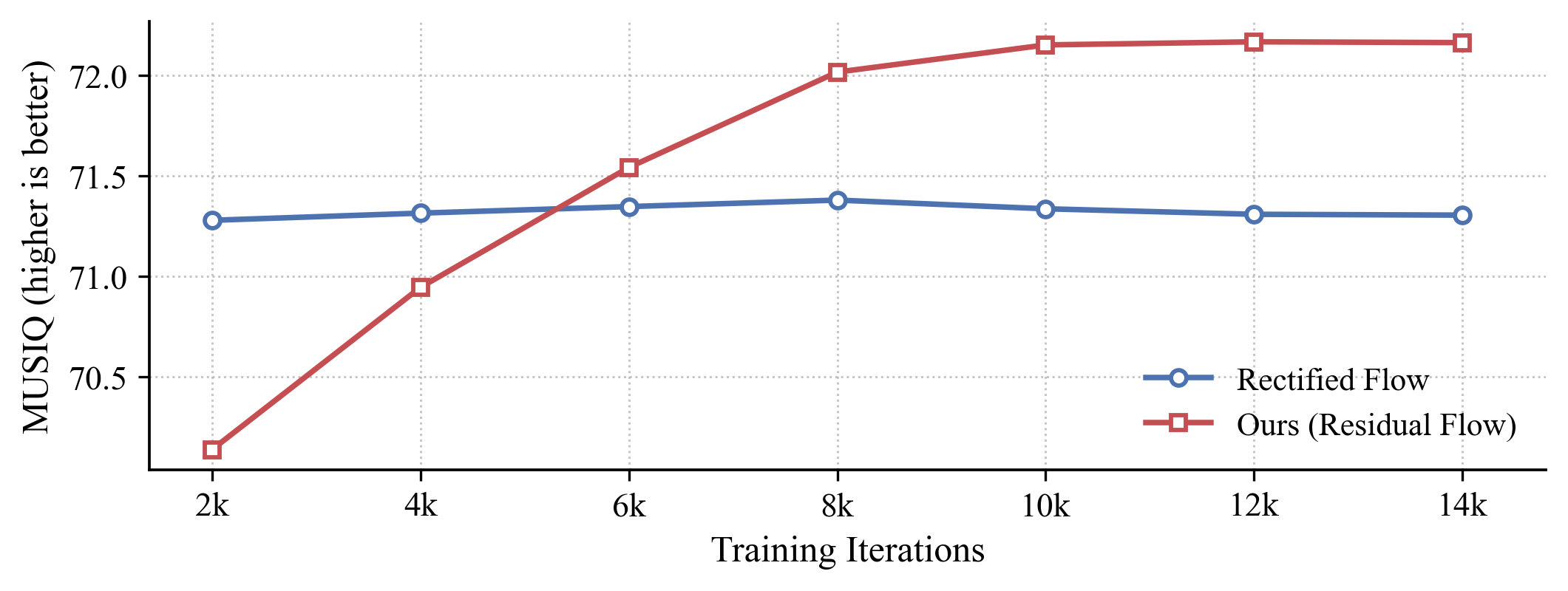}
    \caption{Validation curves of (top) LPIPS and (bottom) MUSIQ with respect to training iterations on LSDIR, evaluated with $15$ inference steps.}
    \label{fig:training_curves}
\end{figure}

In this section, we analyze the design choices of the proposed method and evaluate its performance. 
To simplify the presentation, we refer to our method as RFMSR, standing for \textit{Residual Flow Matching for Image Super-Resolution}.

\subsection{Experiment Settings}
\label{sec:settings}

\textbf{Training and Testing Datasets.} To construct a diverse training set, we collect images from public sources including LSDIR~\cite{li2023lsdir}, DIV2K~\cite{agustsson2017ntire}, Flickr2K~\cite{lim2017enhanced}, and FFHQ~\cite{karras2019style}, as well as web images subjected to quality filtering. In total, approximately 1M high-quality images are curated for training. We then synthesize HQ-LQ training pairs using the Real-ESRGAN degradation pipeline~\cite{wang2021real}.

For evaluation, we adopt ScreenSR~\cite{wu2026vosr} and randomly sample 250 images from the LSDIR~\cite{li2023lsdir} and ImageNet~\cite{russakovsky2015imagenet} test splits, respectively. For ScreenSR, both the LQ input and the GT image are of size $512 \times 512$. For LSDIR and ImageNet-Test, all HQ images are randomly cropped to $512 \times 512$ resolution, and the corresponding LQ images have a resolution of $128 \times 128$. The LQ images of ImageNet-Test are synthesized with the Real-ESRGAN degradation pipeline~\cite{wang2021real}.
\\
\textbf{Compared Methods.} We compare RFMSR against representative real-world ISR methods, including multi-step approaches ResShift~\cite{yue2023resshift}, SeeSR~\cite{wu2024seesr}, and VOSR~\cite{wu2026vosr}, as well as single-step methods OSEDiff~\cite{wu2024one}, InvSR~\cite{yue2025arbitrary}, and the single-step distillation variant of VOSR~\cite{wu2026vosr}. 
Among these, SeeSR, OSEDiff, and InvSR are built upon large-scale T2I foundation models, while ResShift and VOSR are vision-only approaches. 
All comparisons follow the official configurations of each method. For VOSR, we evaluate both its 0.5B and 1.4B parameter variants. Similarly, we report RFMSR at both scales, where the multi-step variant is trained using only Phase~I and the single-step variant is obtained after Phase~II.
\\
\textbf{Evaluation Metrics.} We evaluate all methods with full-reference and no-reference metrics. For distortion fidelity, we report PSNR and SSIM~\cite{wang2004image} on the Y channel of the YCbCr color space. For reference-based perceptual quality, we use LPIPS~\cite{zhang2018unreasonable}, DISTS~\cite{ding2020image}, and FID~\cite{heusel2017gans}. 
For no-reference perceptual quality, we report NIQE~\cite{mittal2012making}, MUSIQ~\cite{ke2021musiq}, MANIQA~\cite{yang2022maniqa}, and CLIPIQA~\cite{wang2023exploring}.
\\
\textbf{Model Architecture.} RFMSR adopts the model architecture of VOSR~\cite{wu2026vosr}, the current SOTA vision-only ISR method, at two scales with 0.5B and 1.4B parameters, both built upon the LightningDiT~\cite{yao2025reconstruction} backbone.
The 0.5B variant encodes images with the SD2.1 VAE~\cite{podell2024sdxl} and extracts semantics with DINOv2-Base~\cite{oquab2023dinov2}, whereas the 1.4B variant instead adopts a 16-channel latent VAE---a 2D variant of the Qwen-Image VAE~\cite{wu2025qwenimagetechnicalreport}---together with DINOv2-Large~\cite{oquab2023dinov2} for semantic encoding.
In both cases, the model takes the encoded LQ latent $y_0$ and the DINOv2 semantic features as the joint conditioning input $c = (y_0, \,\text{DINOv2}(\text{LQ Image}))$. In Phase~I, we initialize the model from the pre-trained VOSR weights of the corresponding scale.

For all experiments reported in the main text, the noise level is set to $\sigma = 1.0$. Detailed training hyper-parameters and configurations are provided in the \textbf{Appendix}. All ablation studies are conducted with the 0.5B variant.

\subsection{Comparison with State-of-the-Arts}
\label{sec:sota_comparison}

\textbf{Quantitative Comparisons.} \cref{tab:main_comparison} reports the quantitative results on LSDIR, ImageNet-Test, and ScreenSR. 
In the multi-step setting, RFMSR achieves competitive or superior perceptual quality across all three benchmarks: on LSDIR the 1.4B variant leads in LPIPS, DISTS, MUSIQ, and FID while the 0.5B variant leads in MANIQA and CLIPIQA; on ImageNet-Test RFMSR attains the best DISTS, NIQE, CLIPIQA, and FID, with the 1.4B variant additionally achieving the best MUSIQ; and on ScreenSR it attains the best MUSIQ, MANIQA, CLIPIQA, and FID among multi-step methods. 
Notably, these results are obtained with fewer inference steps than SeeSR and VOSR.
\\
\hspace*{\parindent}In the single-step setting, the two variants occupy the top two positions on LPIPS, DISTS, and FID across all three datasets, and remain competitive on MANIQA and CLIPIQA.
The two scales are complementary, with the 1.4B variant stronger in DISTS and SSIM and the 0.5B variant stronger in PSNR, MANIQA, and FID on ScreenSR.
Consistent gains over the distilled VOSR at the same scale indicate that directly supervising the single-step prediction improves single-step quality without an additional distillation stage.
\\
\textbf{Qualitative Comparisons.} \cref{fig:sr_comparison} provides visual comparisons under both multi-step and single-step settings, where the portrait sample is shown for the 0.5B variants and the text logo for the 1.4B variants.
For the portrait sample, OSEDiff, InvSR and VOSR exhibit local artifacts and unnatural rendering in facial details, while SeeSR yields slightly smoothed textures and ResShift produces blurry results; RFMSR achieves well-balanced texture recovery and structural fidelity.
For the text logo, SeeSR, ResShift, InvSR, OSEDiff and VOSR all produce distorted glyphs with misspelling and edge blurring, whereas RFMSR maintains accurate structural reconstruction.
More visual comparisons are provided in the \textbf{Appendix}.

\subsection{Ablation Studies}
\label{sec:ablation_studies}

\textbf{Effectiveness of Residual Flow.} To validate the gains brought by our residual flow formulation, we compare it against the standard rectified flow~\cite{liu2022flow} under identical training settings. 
Note that, for a fair comparison, both flow matching models are trained without the classifier-free guidance (CFG)~\cite{ho2022classifier} strategy used in VOSR; consequently, the rectified-flow results reported here differ slightly from the original results of VOSR.
The results are reported in \cref{tab:flow_comparison} and \cref{fig:training_curves}.
As shown in \cref{tab:flow_comparison}, the residual flow consistently outperforms the rectified flow across all metrics and step counts, in terms of both distortion fidelity (e.g., PSNR) and perceptual quality (e.g., LPIPS, MUSIQ, CLIPIQA). This advantage persists from low to high step counts, and the gap remains stable across all evaluation settings. 
\cref{fig:training_curves} further plots the validation LPIPS and MUSIQ against training iterations on LSDIR at $15$ inference steps.
Since VOSR, on which our architecture is built, is originally trained with the rectified flow, its rectified-flow performance can be regarded as an already well-saturated baseline.
Training the same architecture with the residual flow further improves the performance, and once both curves saturate, the residual flow is clearly superior to the original rectified flow.
Overall, these results confirm that setting the source distribution to an approximation of the LQ image distribution preserves the structural priors already contained in the LQ image and consistently improves the performance.

\begin{table}[t]
    \centering
    \caption{Multi-step comparison of standard rectified flow and the proposed residual flow on ImageNet-Test.}
    \label{tab:flow_comparison}
    \setlength{\tabcolsep}{4pt}
    \footnotesize
    \begin{tabular*}{\linewidth}{@{\extracolsep{\fill}}l c cccc@{}}
        \toprule
        Flow Type & Steps & PSNR$\uparrow$ & LPIPS$\downarrow$ & MUSIQ$\uparrow$ & CLIPIQA$\uparrow$ \\
        \midrule
        \multirow{4}{*}{Rectified Flow} & 5  & 26.31 & 0.2495 & 68.8500 & 0.7211 \\
                                        & 10 & 25.78 & 0.2422 & 70.4434 & 0.7206 \\
                                        & 15 & 25.44 & 0.2418 & 70.9141 & 0.7165 \\
                                        & 20 & 25.30 & 0.2390 & 71.0753 & 0.7154 \\
        \midrule
        \multirow{4}{*}{Residual Flow} & 5  & 26.61 & 0.2426 & 69.1112 & 0.7235 \\
                                       & 10 & 26.04 & 0.2345 & 70.6894 & 0.7243 \\
                                       & 15 & 25.77 & 0.2330 & 71.1203 & 0.7242 \\
                                       & 20 & 25.63 & 0.2326 & 71.3034 & 0.7249 \\
        \bottomrule
    \end{tabular*}
\end{table}

\noindent\textbf{Effectiveness of Single-Step Training Strategies.} To evaluate the effectiveness of our proposed two-phase training paradigm for single-step generation, we compare Phase~II against consistency learning~\cite{song2023consistency}, as reported in \cref{tab:phase_comparison}. 
The detailed implementation of consistency learning is provided in the \textbf{Appendix}. As shown in the table, Phase~I (pure CFM training) serves as a reference baseline with limited single-step perceptual quality. 
Consistency training achieves competitive single-step performance with strong no-reference perceptual metrics, while its fidelity metrics are somewhat degraded. 
However, consistency distillation also removes the model's multi-step capability.
Our Phase~II, in contrast, substantially improves single-step generation quality while fully retaining the multi-step capability, supporting inference with an arbitrary number of steps.
Compared with the single-step setting, increasing the number of inference steps further improves the no-reference perceptual metrics, whereas single-step inference offers the highest efficiency with competitive fidelity.

\begin{table}[t]
    \centering
    \caption{Comparison of different training strategies at multiple inference step counts on the LSDIR dataset.}
    \label{tab:phase_comparison}
    \setlength{\tabcolsep}{4pt}
    \footnotesize
    \begin{tabular*}{\linewidth}{@{\extracolsep{\fill}}l c cccc@{}}
        \toprule
        Method & Steps & PSNR$\uparrow$ & LPIPS$\downarrow$ & MUSIQ$\uparrow$ & CLIPIQA$\uparrow$ \\
        \midrule
        \multirow{3}{*}{Phase~I}  & 1-step  & 21.00 & 0.2841 & 65.9456 & 0.6737 \\
                                   & 5-step  & 20.48 & 0.2216 & 71.0931 & 0.7469 \\
                                   & 15-step & 20.07 & 0.2127 & 72.1519 & 0.7428 \\
        \midrule
        \multirow{3}{*}{Phase~II} & 1-step  & 20.10 & 0.1961 & 71.7196 & 0.7111 \\
                                   & 5-step  & 20.26 & 0.2115 & 71.7807 & 0.7373 \\
                                   & 15-step & 19.91 & 0.2124 & 72.4163 & 0.7467 \\
        \midrule
        \multirow{3}{*}{Consistency} & 1-step  & 19.93 & 0.2336 & 73.0477 & 0.7594 \\
                                      & 5-step  & 16.97 & 0.4411 & 66.1956 & 0.5557 \\
                                      & 15-step & 16.00 & 0.5544 & 60.9025 & 0.4312 \\
        \bottomrule
    \end{tabular*}
\end{table}

\noindent\textbf{Ablation Study on Phase~II Loss Functions.}
As shown in \cref{tab:loss_ablation}, removing the GAN loss (w/o GAN) yields the highest single-step fidelity, with slightly higher PSNR than the full Phase~II at the expense of lower MUSIQ and CLIPIQA. 
Removing the velocity loss (w/o Velocity), however, leads to consistent degradation as the step count increases: compared with the full Phase~II, it lowers the single-step fidelity metrics while slightly improving MUSIQ, and its multi-step performance deteriorates substantially, with notable drops in both fidelity and perceptual metrics. 
This indicates that the velocity loss is essential for preserving the flow-based generative prior and enabling meaningful multi-step refinement. 
The full Phase~II, which combines the CFM velocity loss with LPIPS and GAN, achieves the best overall trade-off, maintaining strong single-step quality while fully retaining multi-step inference capability.

\begin{table}[t]
    \centering
    \caption{Ablation study on the loss terms in the Phase~II training objective. All variants are initialized from the Phase~I checkpoint. Results are evaluated on the LSDIR dataset.}
    \label{tab:loss_ablation}
    \setlength{\tabcolsep}{4pt}
    \footnotesize
    \begin{tabular*}{\linewidth}{@{\extracolsep{\fill}}l c cccc@{}}
        \toprule
        Method & Steps & PSNR$\uparrow$ & LPIPS$\downarrow$ & MUSIQ$\uparrow$ & CLIPIQA$\uparrow$ \\
        \midrule
        \multirow{3}{*}{w/o GAN}      & 1-step  & 20.28 & 0.1959 & 71.4848 & 0.6858 \\
                                      & 5-step  & 20.33 & 0.2162 & 71.9721 & 0.7291 \\
                                      & 15-step & 19.94 & 0.2155 & 72.6567 & 0.7334 \\
        \midrule
        \multirow{3}{*}{w/o Velocity} & 1-step  & 19.91 & 0.1992 & 72.2882 & 0.7108 \\
                                      & 5-step  & 19.43 & 0.2212 & 71.0794 & 0.7081 \\
                                      & 15-step & 18.99 & 0.2745 & 68.5734 & 0.6686 \\
        \midrule
        \multirow{3}{*}{Phase~II}     & 1-step  & 20.10 & 0.1961 & 71.7196 & 0.7111 \\
                                      & 5-step  & 20.26 & 0.2115 & 71.7807 & 0.7373 \\
                                      & 15-step & 19.91 & 0.2124 & 72.4163 & 0.7467 \\
        \bottomrule
    \end{tabular*}
\end{table}

\noindent\textbf{Ablation Study on Noise Level.} The noise level $\sigma$ controls the initial perturbation strength around the LQ latent $y_0$ in the residual flow formulation~\cref{eq:rfmsr_path}.
We train separate models with $\sigma \in \{0.0, 0.5, 1.0, 2.0\}$ under the same training configuration, and evaluate them on LSDIR using $15$ uniform inference steps.
As shown in \cref{tab:sigma_ablation}, $\sigma = 0.0$ attains the highest PSNR and SSIM, which is expected since removing the noise term reduces the objective to a deterministic $\mathcal{L}_2$ regression (\cref{sec:why_noise}) that favors pixel-wise fidelity at the expense of perceptual quality.
A modest noise level ($\sigma = 0.5$) yields higher PSNR and SSIM than $\sigma = 1.0$, but falls behind on all the other perceptual metrics.
Increasing $\sigma$ to $2.0$ brings a higher MANIQA than $\sigma = 1.0$ while degrading all the remaining metrics.
We therefore adopt $\sigma = 1.0$ as the default, which strikes the best overall balance between fidelity and perceptual quality.
\begin{table}[t]
    \centering
    \caption{Ablation study of the noise level $\sigma$ on the LSDIR dataset. All results are obtained with 15 uniform inference steps.}
    \label{tab:sigma_ablation}
    \setlength{\tabcolsep}{3.5pt}
    \footnotesize
    \begin{tabular*}{\linewidth}{@{\extracolsep{\fill}}c cccccc@{}}
        \toprule
        $\sigma$ & PSNR$\uparrow$ & SSIM$\uparrow$ & LPIPS$\downarrow$ & DISTS$\downarrow$ & MUSIQ$\uparrow$ & MANIQA$\uparrow$ \\
        \midrule
        0.0 & 20.51 & 0.5038 & 0.4142 & 0.2894 & 57.2390 & 0.4070 \\
        0.5 & 20.12 & 0.4925 & 0.2167 & 0.1302 & 71.9759 & 0.5703 \\
        1.0 & 20.07 & 0.4914 & 0.2127 & 0.1293 & 72.1519 & 0.5792 \\
        2.0 & 20.05 & 0.4898 & 0.2136 & 0.1325 & 71.9859 & 0.5887 \\
        \bottomrule
    \end{tabular*}
\end{table}

\noindent\textbf{Additional Studies.} Further experimental results and details are provided in the \textbf{Appendix}.

\section{Conclusion}
\label{sec:conclusion}

In this paper, we propose RFMSR, a flow matching framework that defines the source distribution as a Gaussian centered at the LQ latent $y_0$, \ie, $p_1 = \mathcal{N}(y_0, \sigma^2 I)$, so that the flow starts from an approximation of the LQ image rather than from pure noise.
This residual flow preserves the structural priors contained in the LQ image and consistently improves the quality of the restored images over the standard rectified flow.
To achieve high-quality single-step generation, we further propose a simple yet effective training strategy that combines the supervision loss on the single-step output with the CFM loss, enabling high-quality restoration with an arbitrary number of steps without resorting to distillation or other acceleration techniques. In future work, extending the residual flow matching framework to other image restoration tasks, such as denoising, deblurring, and inpainting, is a promising direction. Moreover, adopting more advanced supervision losses for single-step acceleration to further improve the generation quality is also worth exploring.

\newpage
{
    \small
    \bibliographystyle{ieeenat_fullname}
    \bibliography{main}
}
\clearpage
\appendix
\section{Appendix}
\label{sec:appendix}

In this appendix, we provide the following contents:
\begin{itemize}
    \item Implementation details of RFMSR, including the training hyper-parameters, the complete two-phase training algorithm, and the computational resources;
    \item Implementation details of consistency learning used as a single-step baseline;
    \item Efficiency comparison in terms of parameter counts and inference runtime;
    \item Additional quantitative comparisons on the real-world RealSR and DRealSR benchmarks;
    \item Additional qualitative comparisons and further visual examples of the noise-level ablation.
\end{itemize}

\subsection{Training Details}

A summary of the training hyper-parameters for Phase~I and Phase~II is provided in \cref{tab:training_params}. Detailed descriptions of training configurations are given below.

\begin{table}[h]
    \centering
    \caption{Training hyper-parameters for Phase~I and Phase~II.}
    \label{tab:training_params}
    \small
    \begin{tabularx}{\linewidth}{@{}Xcc@{}}
        \toprule
        Hyper-parameter & Phase~I & Phase~II \\
        \midrule
        Optimizer & AdamW & AdamW \\
        Learning rate & $5\times 10^{-5}$ & $5\times 10^{-5}$ \\
        $\beta_1 / \beta_2$ & 0.9 / 0.999 & 0.9 / 0.999 \\
        Weight decay & $0.0$ & $0.0$ \\
        Learning rate schedule & - & - \\
        Gradient clipping & 1.0 & 1.0 \\
        Batch size & 64 & 32 \\
        Training iterations & 20k & 20k \\
        EMA decay & 0.999 & 0.999 \\
        Timestep distribution & uniform & uniform \\
        LQ size & $128 \times 128$ & $128 \times 128$ \\
        HQ size & $512 \times 512$ & $512 \times 512$ \\
        Noise level $\sigma$ & 1.0 & 1.0 \\
        $\mathcal{L}(\theta)$ & \checkmark & \checkmark \\
        $\lambda_{\text{LPIPS}}$ & - & 1.0 \\
        $\lambda_{\text{GAN}}$ & - & 0.1 \\
        \bottomrule
    \end{tabularx}
\end{table}

\noindent\textbf{Phase I.}
We train the velocity field $v_\theta$ using the AdamW optimizer \cite{loshchilov2017decoupled} with $\beta_1 = 0.9$, $\beta_2 = 0.999$, and no weight decay.
The learning rate is set to $5\times 10^{-5}$ without a learning rate schedule. We use a batch size of $64$ and train for $20$k iterations.
The noise level $\sigma$ is set to $1.0$. HQ images are randomly cropped to $512\times 512$, and the corresponding $128\times 128$ LQ images are synthesized using the Real-ESRGAN degradation pipeline~\cite{wang2021real}.

\noindent\textbf{Phase II.}
We fine-tune the Phase~I checkpoint for an additional $20$k iterations, using the same optimizer settings with a reduced batch size of $32$.
The LPIPS loss weight $\lambda_{\text{LPIPS}}$ is set to $1.0$ and the GAN loss weight $\lambda_{\text{GAN}}$ is set to $0.1$.
The discriminator follows the PatchGAN~\cite{isola2017image} architecture with hinge loss~\cite{miyato2018spectral}.
The Phase~II training procedure is summarized in Algorithm~\ref{alg:two_phase}.

\noindent\textbf{Computational Resources.}
All training and inference experiments reported in this paper are conducted on a single NVIDIA RTX PRO 6000 GPU with 96GB of memory.

\subsection{Alternative Single-Step Training Strategies}
\label{sec:A_consistency_reflow}

We implement consistency learning~\cite{song2023consistency} as a representative single-step acceleration baseline for comparison with the proposed Phase~II. It shares the same backbone architecture and residual flow formulation as RFMSR, and is initialized from the Phase~I checkpoint.

\medskip
\noindent\textbf{Consistency Learning.}
Following Consistency Models~\cite{song2023consistency}, we define a consistency function $f_\theta$ that directly maps any point on the residual flow trajectory to the clean HQ latent $x_0$:
\begin{equation}
    f_\theta(x_t, t) = x_t - t \cdot v_\theta(x_t, t, c),
\end{equation}
which satisfies the boundary condition $f_\theta(x, 0) = x$. The flow path follows the same formulation as RFMSR: $x_t = x_0 + t\,(y_0 - x_0) + t\,\sigma\,\varepsilon$.
During training, we sample a random timestep $t \in [t_{\text{min}}, 1]$ and construct two adjacent trajectory points $x_t$ and $x_{t - \Delta t}$ using the same noise $\varepsilon$ to ensure trajectory consistency. The online model predicts $f_\theta(x_t, t)$, while a target model (an EMA of the online model with decay rate $0.999$) predicts $f_{\theta^-}(x_{t - \Delta t},\, t - \Delta t)$. The consistency training loss enforces agreement between the two predictions:
\begin{equation}
    \mathcal{L}_{\text{CT}} = \mathbb{E}_{t, \varepsilon} \big[ \| f_\theta(x_t, t) - \text{sg}[f_{\theta^-}(x_{t - \Delta t},\, t - \Delta t)] \|^2 \big],
\end{equation}
where $\text{sg}$ denotes stop-gradient. We set $t_{\text{min}} = 10^{-3}$ to avoid numerical instability at $t = 0$, and $\Delta t = 0.05$. The target network is updated after each optimizer step via EMA, and the training runs for $30$k iterations with the same optimizer configuration.

\subsection{Efficiency Comparison}
\label{sec:A_efficiency}
\begin{table*}[t]
    \centering
    \caption{Efficiency comparison on $512 \times 512$ inputs. Runtime is measured on an NVIDIA RTX PRO 6000 GPU with batch size 1 in FP16.}
    \label{tab:efficiency}
    \resizebox{\linewidth}{!}{%
    \begin{tabular}{@{}l cccccc cccccc@{}}
        \toprule
        \multirow{2}{*}{Metrics} & \multicolumn{6}{c}{Multi-Steps} & \multicolumn{6}{c}{Single-Step} \\
        \cmidrule(lr){2-7}\cmidrule(lr){8-13}
         & SeeSR & VOSR-0.5B & VOSR-1.4B & ResShift & \textbf{RFMSR-0.5B} & \textbf{RFMSR-1.4B} & VOSR-0.5B & VOSR-1.4B & InvSR & OSEDiff & \textbf{RFMSR-0.5B} & \textbf{RFMSR-1.4B} \\
        \midrule
        \#Params (M) & 2550.03 & 655.86 & 1743.04 & 173.92 & \textbf{655.86} & \textbf{1743.04} & 655.86 & 1743.04 & 1323.79 & 1805.10 & \textbf{655.86} & \textbf{1743.04} \\
        Runtime (ms) & 2442 & 759 & 1745 & 331 & \textbf{453} & \textbf{757} & 58 & 59 & 61 & 64 & \textbf{58} & \textbf{59} \\
        \bottomrule
    \end{tabular}%
    }
\end{table*}

\Cref{tab:efficiency} presents the parameter counts and inference runtime of the model components of each method. Among the multi-step methods, the efficiency gain of RFMSR over VOSR stems from the reduced number of inference steps as well as the removal of the classifier-free guidance (CFG)~\cite{ho2022classifier} inference adopted in VOSR.
In the single-step setting, RFMSR and VOSR share the same runtime, since they are built upon the same model backbone.
More detailed efficiency analyses can be found in VOSR~\cite{wu2026vosr}.

\subsection{Additional Quantitative Comparisons}
\label{sec:A_realworld}
\begin{table*}[!t]
    \centering
    \caption{Additional quantitative comparison on RealSR and DRealSR. The best and second-best results are highlighted in \textcolor{red}{red} and \textcolor{blue}{blue}.}
    \label{tab:realworld}
    \footnotesize
    \begin{tabular*}{\linewidth}{@{\extracolsep{\fill}}l l ccccccccc@{}}
        \toprule
        Dataset & Method & PSNR$\uparrow$ & SSIM$\uparrow$ & LPIPS$\downarrow$ & DISTS$\downarrow$ & NIQE$\downarrow$ & MUSIQ$\uparrow$ & MANIQA$\uparrow$ & CLIPIQA$\uparrow$ & FID$\downarrow$ \\
        \midrule
        \multirow{6}{*}{RealSR}  & InvSR-1       & 24.45 & 0.7139 & 0.2852 & 0.2115 & 5.6400 & 68.3500 & 0.4535 & \textbf{\textcolor{red}{0.6779}} & 137.39 \\
                                 & OSEDiff-1     & 25.15 & 0.7338 & 0.2920 & 0.2129 & 5.6402 & 69.0777 & 0.4710 & \textbf{\textcolor{blue}{0.6686}} & 123.54 \\
                                 & VOSR-1-0.5B   & 24.85 & 0.7096 & 0.2904 & 0.2107 & \textbf{\textcolor{red}{5.1440}} & \textbf{\textcolor{blue}{69.9123}} & \textbf{\textcolor{red}{0.4977}} & 0.6017 & 122.46 \\
                                 & VOSR-1-1.4B   & 24.55 & 0.7023 & 0.2869 & 0.2127 & \textbf{\textcolor{blue}{5.4815}} & \textbf{\textcolor{red}{71.0555}} & \textbf{\textcolor{blue}{0.4859}} & 0.6032 & 113.07 \\
                                 & \textbf{RFMSR-1-0.5B} & \textbf{\textcolor{red}{25.93}} & \textbf{\textcolor{red}{0.7577}} & \textbf{\textcolor{blue}{0.2628}} & \textbf{\textcolor{blue}{0.1910}} & 5.7796 & 64.7956 & 0.4565 & 0.6196 & \textbf{\textcolor{blue}{106.88}} \\
                                 & \textbf{RFMSR-1-1.4B} & \textbf{\textcolor{blue}{25.28}} & \textbf{\textcolor{blue}{0.7503}} & \textbf{\textcolor{red}{0.2380}} & \textbf{\textcolor{red}{0.1879}} & 7.7897 & 63.5841 & 0.4097 & 0.5917 & \textbf{\textcolor{red}{106.62}} \\
        \midrule
        \multirow{6}{*}{DRealSR} & InvSR-1       & 26.40 & 0.7153 & 0.3515 & 0.2436 & \textbf{\textcolor{blue}{5.9604}} & 65.8092 & 0.4640 & \textbf{\textcolor{red}{0.7154}} & 167.79 \\
                                 & OSEDiff-1     & \textbf{\textcolor{blue}{27.92}} & \textbf{\textcolor{red}{0.7833}} & 0.2968 & 0.2164 & 6.4397 & 64.6918 & \textbf{\textcolor{blue}{0.4656}} & \textbf{\textcolor{blue}{0.6964}} & 135.52 \\
                                 & VOSR-1-0.5B   & 27.31 & 0.7343 & 0.3543 & 0.2418 & \textbf{\textcolor{red}{5.8347}} & \textbf{\textcolor{blue}{66.0519}} & \textbf{\textcolor{red}{0.4727}} & 0.6133 & 169.92 \\
                                 & VOSR-1-1.4B   & 27.08 & 0.7152 & 0.3457 & 0.2381 & 6.0234 & \textbf{\textcolor{red}{66.3979}} & 0.4497 & 0.6220 & 163.51 \\
                                 & \textbf{RFMSR-1-0.5B} & \textbf{\textcolor{red}{28.53}} & \textbf{\textcolor{red}{0.7833}} & \textbf{\textcolor{blue}{0.2963}} & \textbf{\textcolor{red}{0.2068}} & 6.2171 & 57.8906 & 0.4036 & 0.6258 & \textbf{\textcolor{red}{131.17}} \\
                                 & \textbf{RFMSR-1-1.4B} & 27.53 & \textbf{\textcolor{blue}{0.7621}} & \textbf{\textcolor{red}{0.2830}} & \textbf{\textcolor{blue}{0.2100}} & 8.4569 & 56.0722 & 0.3612 & 0.5994 & \textbf{\textcolor{blue}{133.45}} \\
        \bottomrule
    \end{tabular*}
\end{table*}
\begin{table}[t]
    \centering
    \caption{No-reference metrics of the HQ images in RealSR, DRealSR, and ScreenSR, which provide the reference values for the no-reference perceptual metrics.}
    \label{tab:hq_reference}
    \setlength{\tabcolsep}{4pt}
    \footnotesize
    \begin{tabular*}{\linewidth}{@{\extracolsep{\fill}}l cccc@{}}
        \toprule
        Dataset & NIQE$\downarrow$ & MUSIQ$\uparrow$ & MANIQA$\uparrow$ & CLIPIQA$\uparrow$ \\
        \midrule
        RealSR  & 6.1167 & 57.4614 & 0.3295 & 0.4593 \\
        DRealSR & 6.7900 & 50.5643 & 0.3152 & 0.4756 \\
        ScreenSR & 3.7716 & 72.1500 & 0.5651 & 0.6866 \\
        \bottomrule
    \end{tabular*}
\end{table}

We compare the single-step variants of RFMSR with several state-of-the-art ISR methods on two real-world datasets, RealSR~\cite{cai2019toward} and DRealSR~\cite{wei2020component}, using the cropped LQ-HQ pairs from OSEDiff~\cite{wu2024one} with sizes of $128 \times 128$ and $512 \times 512$, respectively.
As reported in \cref{tab:realworld}, RFMSR achieves the best PSNR, SSIM, LPIPS, DISTS, and FID on both benchmarks, whereas its no-reference scores are relatively lower than those of some competitors.
\cref{tab:hq_reference} lists the no-reference metrics of the HQ images.
On RealSR and DRealSR, the HQ images themselves obtain low no-reference scores, and the no-reference scores of RFMSR are also relatively low on these two benchmarks.
In contrast, the HQ images of ScreenSR obtain considerably higher no-reference scores, and RFMSR correspondingly achieves high no-reference scores on ScreenSR, as reported in \cref{tab:main_comparison}.

\subsection{Additional Visual Comparisons}
\label{sec:A_visual}

\begin{figure*}[t]
    \centering
    \includegraphics[width=\linewidth]{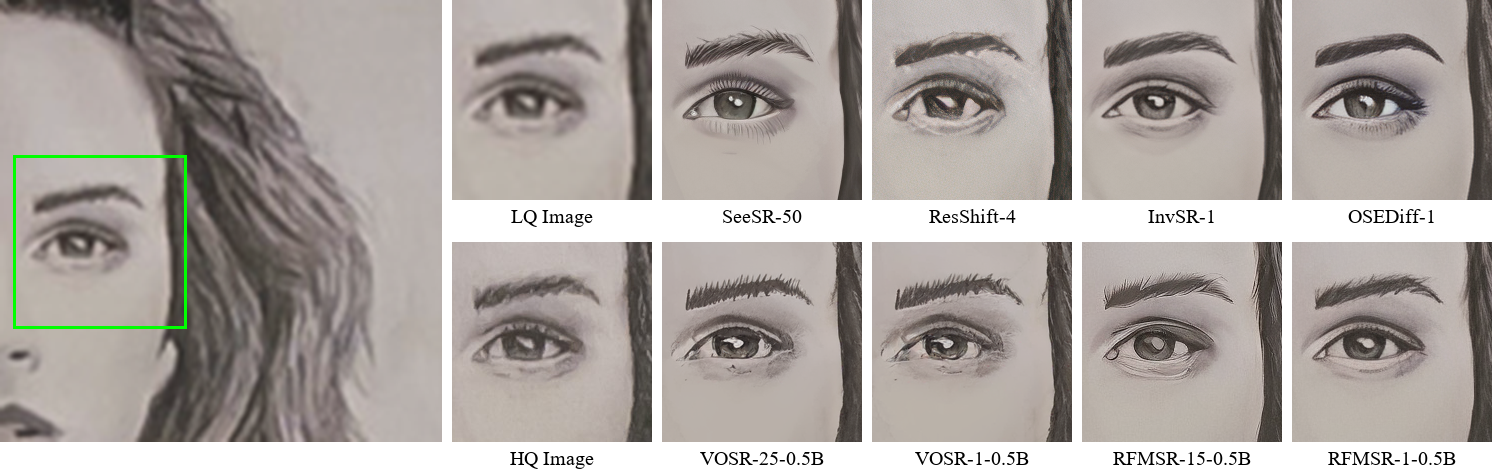} \\[0.5pt]
    \includegraphics[width=\linewidth]{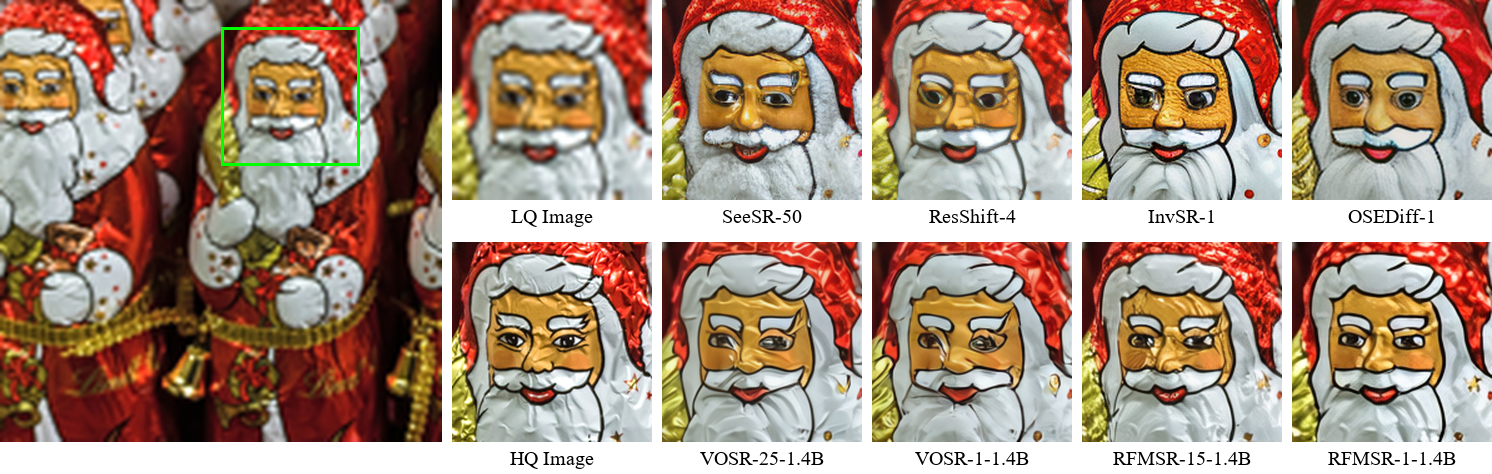} \\[0.5pt]
    \includegraphics[width=\linewidth]{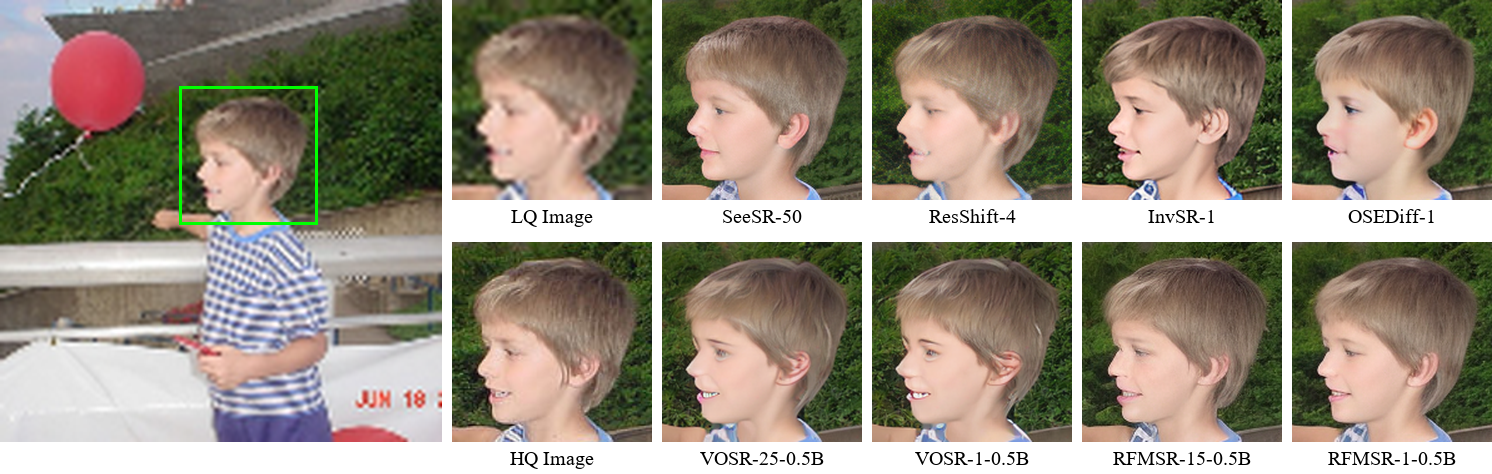}
    \caption{Additional visual comparisons of RFMSR with state-of-the-art methods on real-world images. (Zoom in for best view.)}
    \label{fig:sr_comparison_appendix_1}
\end{figure*}

\begin{figure*}[t]
    \centering
    \includegraphics[width=\linewidth]{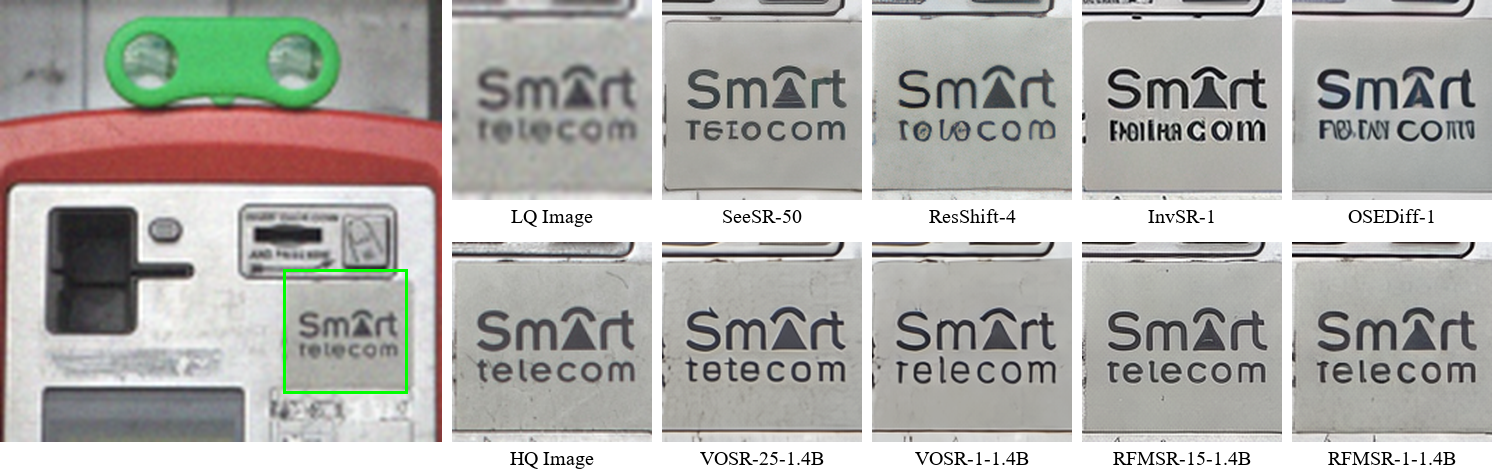} \\[0.5pt]
    \includegraphics[width=\linewidth]{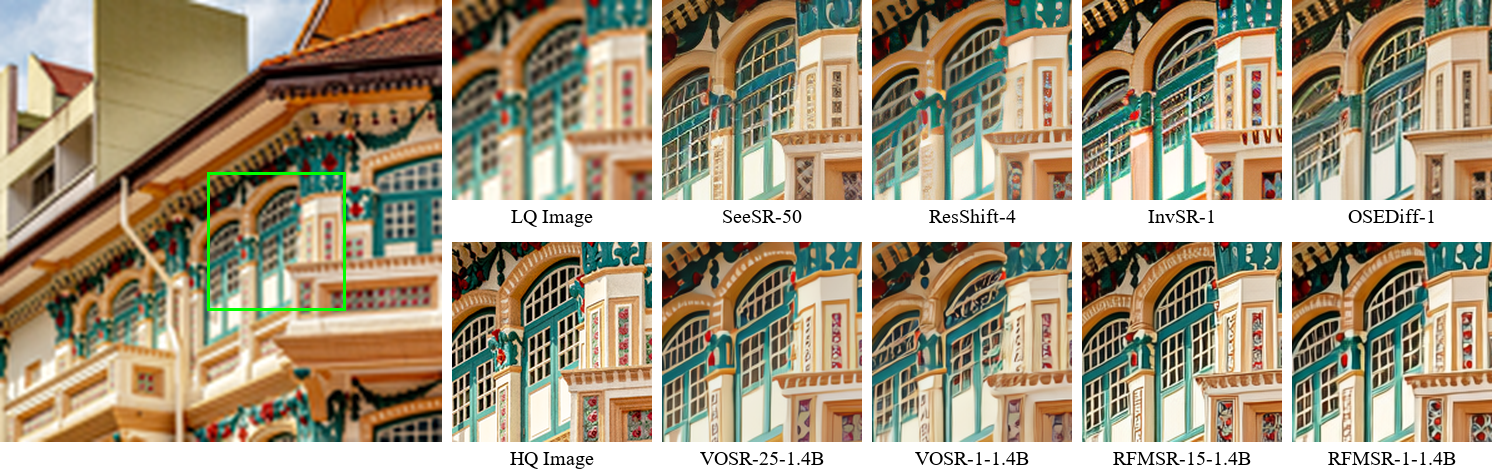} \\[0.5pt]
    \includegraphics[width=\linewidth]{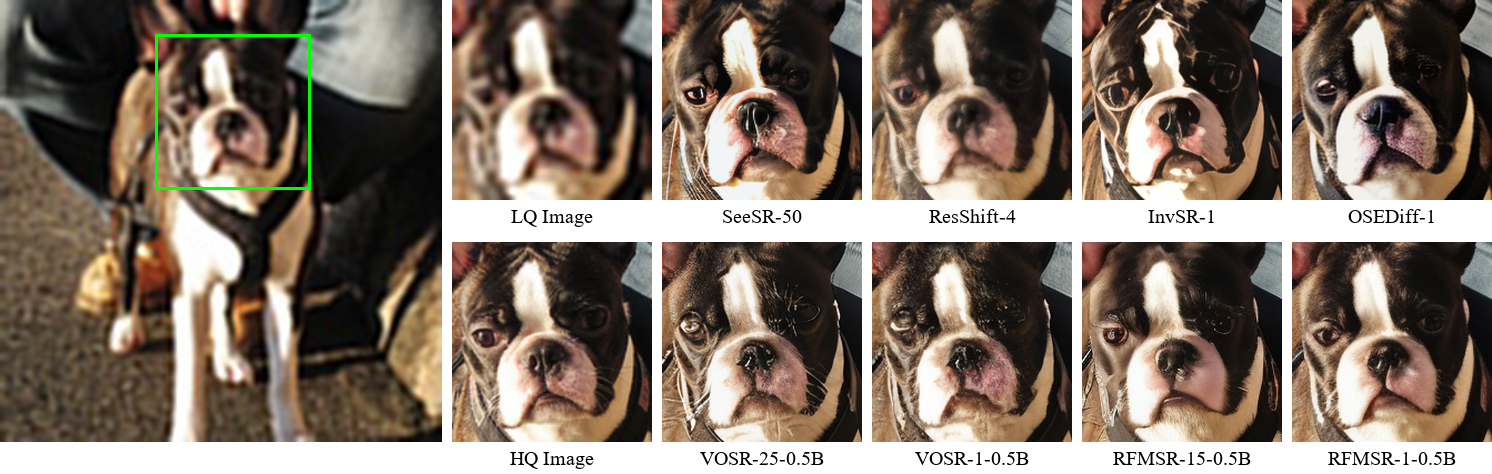}
    \caption{Additional visual comparisons of RFMSR with state-of-the-art methods on real-world images. (Zoom in for best view.)}
    \label{fig:sr_comparison_appendix_2}
\end{figure*}

As shown in \cref{fig:sr_comparison_appendix_1} and \cref{fig:sr_comparison_appendix_2}, the additional visual comparisons cover a diverse range of scenarios, including facial close-ups, decorative objects, outdoor portraits, text signage, architectural structures, and animal fur; each example compares either the 0.5B or the 1.4B variants, as indicated by the method labels.
Across all examples, RFMSR consistently produces the most natural texture and the fewest artifacts among multi-step methods: it recovers eyelashes and eyebrows in the facial close-up without ringing, keeps the fluffy edges of the ornament and the fur of the dog, and reproduces the window grilles and decorative wall panels of the building with faithful geometry.
SeeSR tends to generate texture structures that are inconsistent with the original input, and ResShift often produces over-smoothed, blurry results accompanied by artifacts, whereas RFMSR preserves authentic structural fidelity while recovering rich textures.
In the single-step setting, RFMSR remains close to its multi-step counterpart, whereas VOSR, InvSR, and OSEDiff frequently introduce local artifacts and distorted structures.
For text content, SeeSR, ResShift, InvSR, and OSEDiff alter the glyph shapes of the ``Smart telecom'' sign, whereas VOSR and RFMSR reconstruct the letters correctly.
On the building example, RFMSR also reproduces the fine window grilles and wall ornaments, whereas the other methods tend to generate erroneous and fragmented structures.
These results indicate that the residual flow formulation combined with two-phase training provides a robust foundation for both high-quality single-step generation and progressive multi-step refinement.

\begin{figure*}[t]
    \centering
    \includegraphics[width=\linewidth]{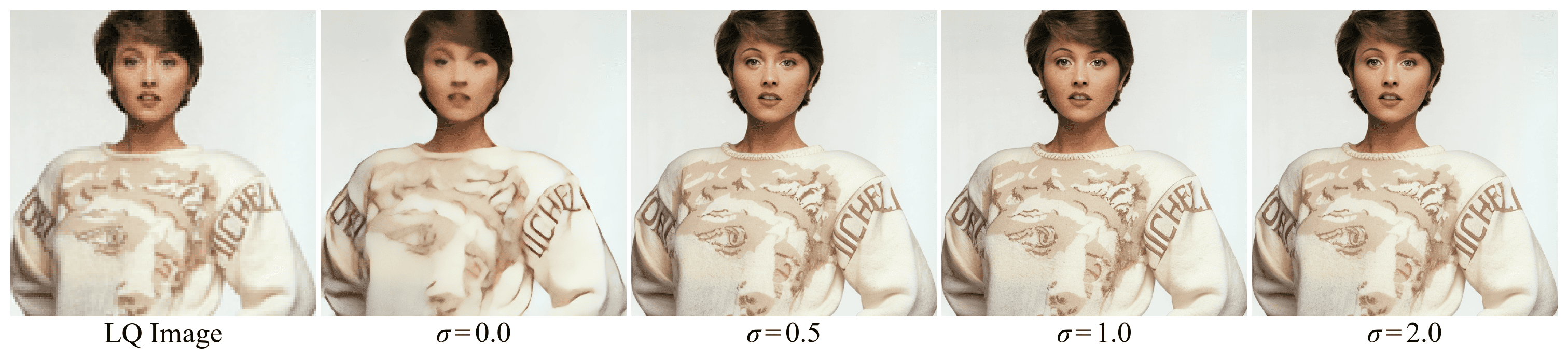} \\[2pt]
    \includegraphics[width=\linewidth]{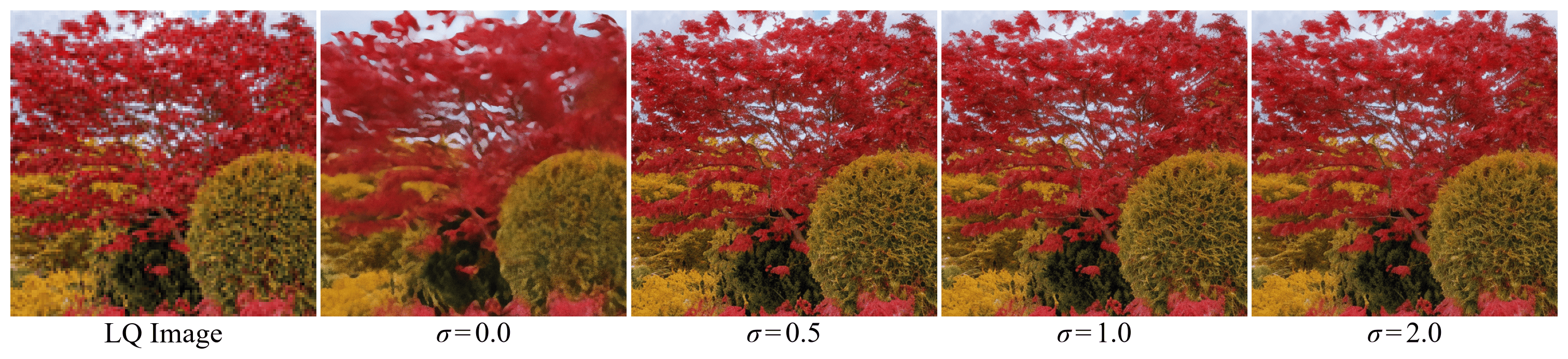} \\[2pt]
    \includegraphics[width=\linewidth]{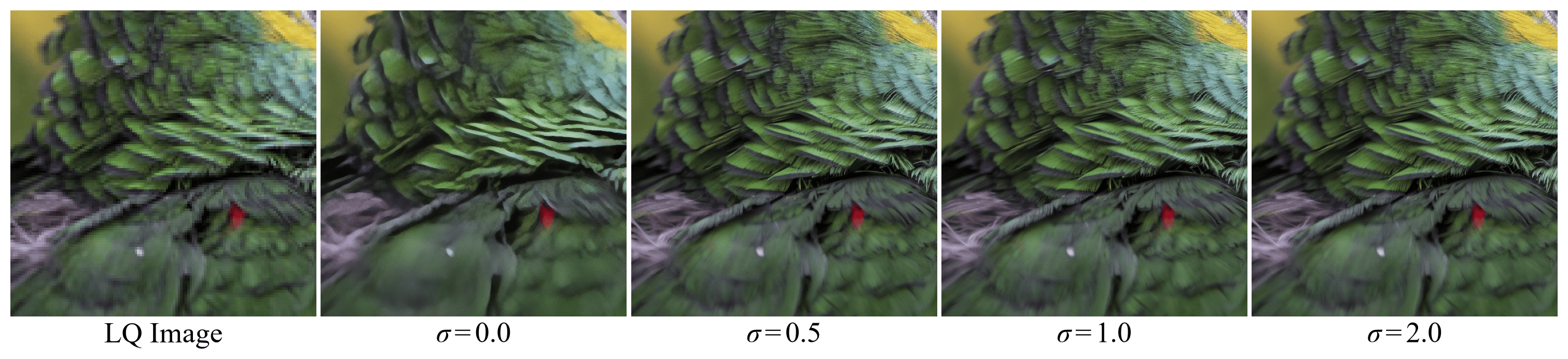} \\[2pt]
    \includegraphics[width=\linewidth]{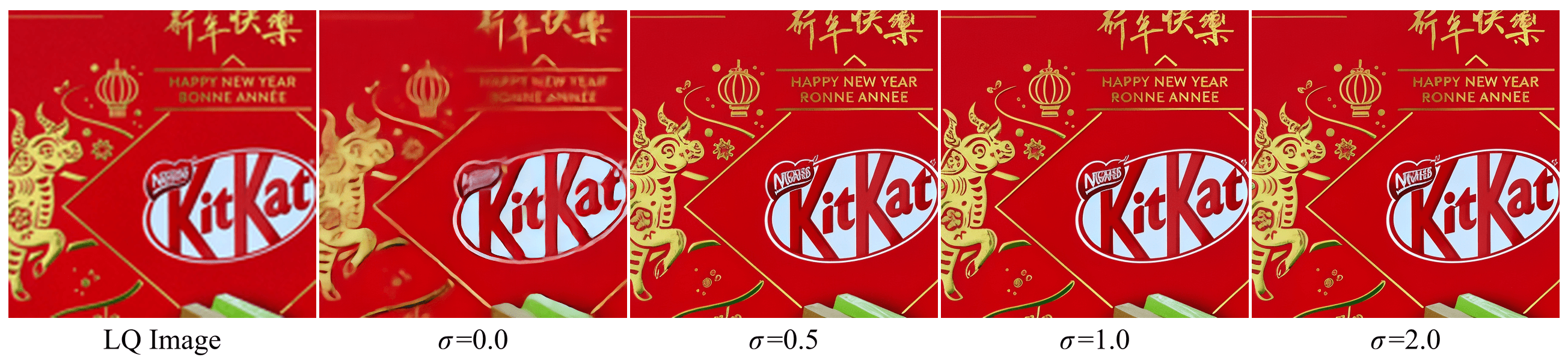} \\[2pt]
    \includegraphics[width=\linewidth]{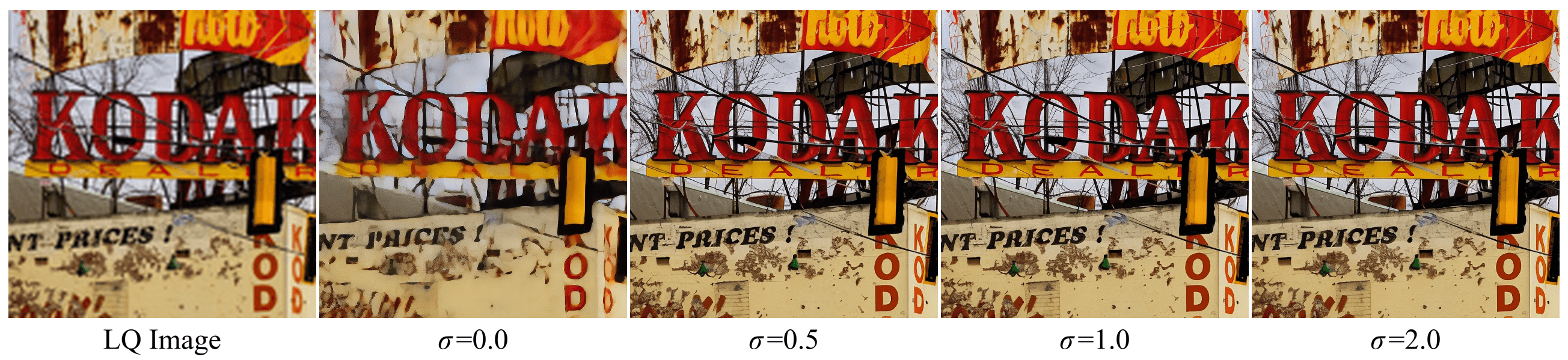}
    \caption{Additional visual examples of the noise level $\sigma$ ablation. (Zoom in for best view.)}
    \label{fig:sigma_ablation_appendix}
\end{figure*}

In addition, \cref{fig:sigma_ablation_appendix} provides further visual examples of the noise-level ablation, showing that the deterministic setting ($\sigma = 0.0$) produces clearly degraded textures, whereas the stochastic settings recover more realistic details.

\end{document}